\documentclass[11pt]{article}
\usepackage[final]{acl}
\usepackage{times}
\usepackage{latexsym}
\usepackage[T1]{fontenc}
\usepackage[utf8]{inputenc}
\usepackage{microtype}
\usepackage{courier}
\usepackage{textcomp}
\usepackage{upquote}
\usepackage{graphicx}
\usepackage{booktabs}
\usepackage{float}
\usepackage{tabularx}
\usepackage{amsmath}
\usepackage{xspace}
\usepackage{listings}
\usepackage{multicol} % Two-column appendix prose around full-width tables.
\usepackage{longtable} % Captioned prompt tables with continued-page captions.

\newcommand{\sql}[1]{\texttt{#1}}

\newcommand{\fitcol}[1]{\resizebox{\ifdim\width>\columnwidth\columnwidth\else\width\fi}{!}{#1}}
\title{Self-Refining Prompt Induction for Streaming Text-to-SQL}
\author{\textbf{Gaetano Rossiello}$^{1}$, \textbf{Alfio Gliozzo}$^{1}$, \textbf{Nandana Mihindukulasooriya}$^{1}$,\\
        \textbf{Erick Lee}$^{2}$, \textbf{Nabila Siregar}$^{1}$, \textbf{Lisa Amini}$^{1}$ \\
        {\mdseries $^{1}$IBM \quad $^{2}$Confluent}}

\hypersetup{pdftitle={Self-Refining Prompt Induction for Streaming Text-to-SQL},pdfauthor={}}
\begin{document}
\maketitle

\begin{abstract}
Text-to-SQL assistants for streaming engines must handle dialect-specific types, windowing rules, and temporal semantics, including constructs poorly handled by generic prompts. We present Self-Refining Prompt Induction (SRPI), an offline method that converts generation failures into reusable dialect instructions while keeping model weights fixed. A Refiner revises the Generator's prompt using reference-conditioned feedback from three LLM judges; a sampled regression guard checks earlier successes, and a separate Compressor reduces prompt length. Inference uses the frozen prompt without retrieval or query repair. We evaluate cross-domain and cross-model transfer on Confluent Cloud Flink SQL using a controlled synthetic workload seeded by 30 expert-written examples, with 335 induction examples and 257 questions from held-out domains. The same induced prompt improves mean judge scores across seven Generators by 7.1--19.2 percentage points, including an increase from 85.3\% to 97.5\% for the induction model. A separate compression study reduces prompt length by 52.7\% with a 0.19-percentage-point score decrease. An assistant using the induced prompt is in early access, with initial user feedback to inform subsequent refinement.
\end{abstract}

\section{Introduction}
Text-to-SQL assistants broaden access to data platforms, but success on Spider and BIRD does not establish reliability on a particular production engine \citep{yu-etal-2018-spider,li-etal-2023-bird}. Cross-dialect evaluation exposes backend differences \citep{perlitz-etal-2026-polysql}, while enterprise workloads add schema and domain complexity \citep{chen-etal-2026-beaver}.

\begin{figure}[t]
\centering
\includegraphics[width=\columnwidth]{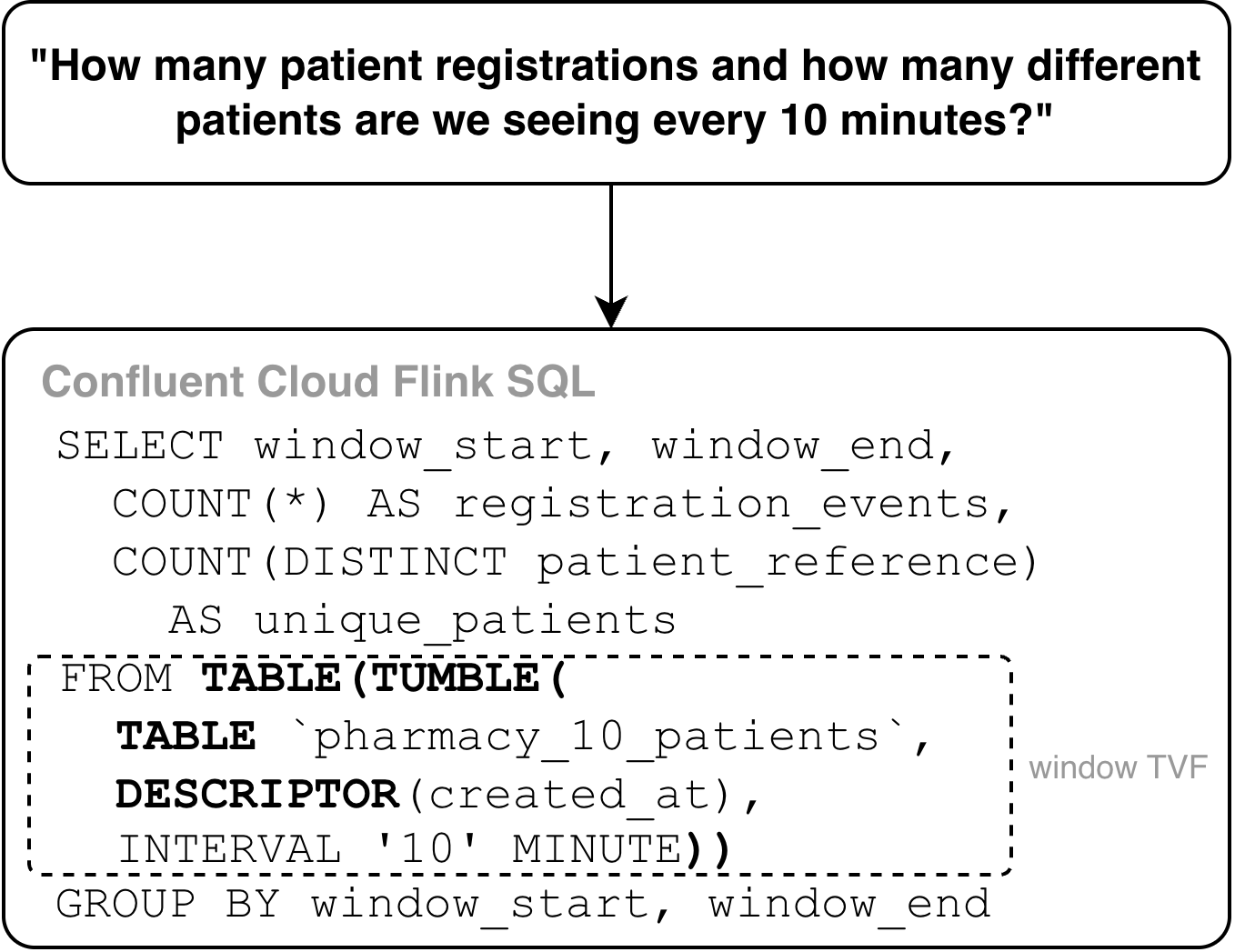}
\caption{A natural-language request and a Flink SQL query over a stream. The dashed region highlights the window table-valued function (TVF), which assigns events to ten-minute windows.}
\label{fig:example}
\end{figure}

% Queue the workflow for the top of page 2.
\begin{figure*}[t]
\centering
\includegraphics[width=\linewidth]{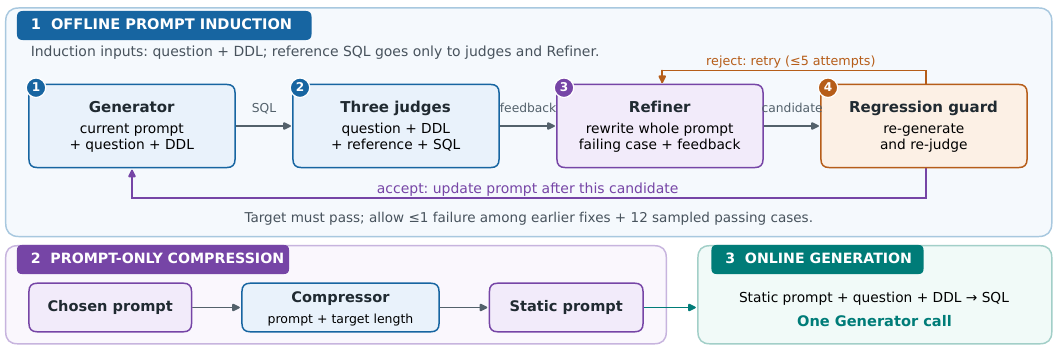}
\caption{SRPI separates offline induction and compression from online generation. Reference SQL is visible to the judges and Refiner, never to the Generator. The guard tests each rewrite before commitment; the requested compression budget need not be met. Online use requires one Generator call with the static prompt.}
\label{fig:pipeline}
\end{figure*}

Our assistant generates Confluent Cloud Flink SQL from English requests and definitions of user-selected tables (Figure~\ref{fig:example}). Streaming SQL unifies operations over streams and tables \citep{begoli-etal-2019-one}, but correct queries still depend on timestamp types, event-time semantics, window membership, and engine-specific conventions. For example, the managed service exposes the Kafka record timestamp through \sql{\$rowtime}; confusing it with a business event timestamp can change a query's meaning \citep{confluent-2026-reference}. Errors can reject a statement or silently change its answer.

Supervised fine-tuning (SFT) adapts weights for specialised Text-to-SQL tasks \citep{tritto-etal-2026-gradesql,glass-etal-2025-extractive}. We instead keep Generator weights fixed and learn reusable instructions. Prompts can be inspected, versioned, and rolled back across Generators without model-specific retraining. This suits inference-only access and frequent model changes, at the cost of longer inputs. Comparison with SFT remains open.

\emph{Self-Refining Prompt Induction} (SRPI) targets a practical maintenance problem: converting failures on long-tail dialect constructs into guidance reusable across domains and Generators. The aim is to encode the target engine's syntax and usage constraints in a prompt. SRPI rewrites instructions using three judges' reference-conditioned feedback, checks earlier successes, and compresses the result. The Generator and Refiner use the same model; judges and optimisation procedure remain fixed.

We study this process in a controlled synthetic workload executed on a live managed cluster. Judge-score targets are 95\% for single-construct questions and 75\% for questions combining at least three constructs. The assistant is in early access, with initial real-user feedback forthcoming. Informal feedback from Flink and streaming-data experts has been encouraging.

The contributions are:
\begin{itemize}
  \item \textbf{System design}: an offline adaptation workflow integrating reference-based feedback, whole-prompt self-refinement, a regression guard, and separate compression.
  \item \textbf{Data and transfer}: a controlled workload testing reuse of one induced prompt across seven Generators and held-out domains within a shared dialect repertoire.
  \item \textbf{Engineering evidence}: measured trade-offs in prompt growth, regression checks, compression, and feedback sources.
\end{itemize}

% Queue the main results for the top of page 3.
\begin{table*}[t]
\centering\fontsize{10}{12}\selectfont\setlength{\tabcolsep}{4.2pt}
% Seed/SRPI pairs and pp gains from reported one-decimal aggregates.
% Bold highlights positive within-Generator gains, not significance.
\begin{tabular}{@{}lrrrrrrrrrrrr@{}}
\toprule
& \multicolumn{3}{c}{Micro} & \multicolumn{3}{c}{Tier 1 ($n{=}191$)} & \multicolumn{3}{c}{Tier 3 ($n{=}22$)} & \multicolumn{3}{c}{U-Yes}\\
\cmidrule(lr){2-4}\cmidrule(lr){5-7}\cmidrule(lr){8-10}\cmidrule(lr){11-13}
Generator & Seed & SRPI & $\Delta$ & Seed & SRPI & $\Delta$ & Seed & SRPI & $\Delta$ & Seed & SRPI & $\Delta$\\
\midrule
GPT-6.1 Sol & 85.3 & 97.5 & \textbf{+12.2} & 87.2 & 97.3 & \textbf{+10.1} & 68.9 & 100.0 & \textbf{+31.1} & 72.0 & 95.7 & \textbf{+23.7} \\
\midrule
GPT-6 Astra & 86.2 & 97.4 & \textbf{+11.2} & 88.3 & 96.9 & \textbf{+8.6} & 69.7 & 100.0 & \textbf{+30.3} & 75.1 & 94.9 & \textbf{+19.8} \\
Claude Opus 5.5 & 85.3 & 97.1 & \textbf{+11.8} & 88.8 & 97.5 & \textbf{+8.7} & 69.7 & 94.7 & \textbf{+25.0} & 72.4 & 93.8 & \textbf{+21.4} \\
Gemini 3.8 Flash & 82.3 & 95.8 & \textbf{+13.5} & 85.3 & 96.5 & \textbf{+11.2} & 69.7 & 90.9 & \textbf{+21.2} & 71.2 & 92.2 & \textbf{+21.0} \\
DeepSeek 4.1 Flash & 70.0 & 89.2 & \textbf{+19.2} & 72.6 & 89.9 & \textbf{+17.3} & 62.9 & 87.9 & \textbf{+25.0} & 57.2 & 81.7 & \textbf{+24.5} \\
Qwen3.8 27B & 61.3 & 77.6 & \textbf{+16.3} & 66.1 & 77.3 & \textbf{+11.2} & 31.1 & 84.8 & \textbf{+53.7} & 48.2 & 72.8 & \textbf{+24.6} \\
GPT-OSS 120B & 46.4 & 53.5 & \textbf{+7.1} & 49.7 & 53.4 & \textbf{+3.7} & 13.6 & 39.4 & \textbf{+25.8} & 35.8 & 45.5 & \textbf{+9.7} \\
\bottomrule
\end{tabular}

\caption{Domain-held-out evaluation ($n=257$), one scored output per question and condition. Scores are percentages; U-Yes is unanimous Yes. $\Delta$ (pp) is SRPI minus Seed from the reported rounded scores. \textbf{Bold}: within-Generator gains. The same 59,787-character SRPI-60k prompt is used throughout; only Sol was used for induction.}
\label{tab:main}
\end{table*}

\section{Workload and Evaluation Data}
\label{sec:data}
SRPI needs questions, table definitions (DDL), and reference SQL. These can be annotated by experts or generated synthetically; the refinement loop accepts either source. Expert annotation requires dialect expertise. We evaluate the synthetic pathway, seeded by 30 expert-written question--SQL pairs validated on a live cluster.

An LLM proposes domain specifications, which are compiled into Flink DDL. The portfolio covers 50 schemas, 45 business domains, 304 tables, and 1,752 columns. Populated tables undergo live integrity checks. The expert examples and official documentation yield 42 SQL patterns in 14 groups, covering windows, joins, pattern matching, JSON operations, and sink statements. Pattern instantiations are executed on the cluster; failures are discarded. An LLM then writes questions from accepted SQL. This checks engine acceptance, not every question--SQL alignment. Evaluated questions explicitly name tables and columns.

The pool contains 3,940 examples with 31 construct labels. Tier~1 has one label (2,864 examples), Tier~2 two (639), and Tier~3 at least three (437). Holding out education, insurance, and retail yields 257 evaluation questions: 191, 44, and 22 by tier, covering 29 labels. The remaining 3,683 form the training pool. Domains and exact SQL are disjoint while pattern templates are shared, directly testing reuse of dialect guidance across domains within the same construct repertoire.

The induction set comprises 335 synthetic examples stratified by domain and construct label. The Refiner sees no evaluation examples. Appendix~\ref{app:data} details generation and coverage.

\section{Self-Refining Prompt Induction}
\label{sec:method}
Figure~\ref{fig:pipeline} shows three stages. \emph{Induction} starts with a 337-character seed prompt and turns errors on training examples into reusable instructions. \emph{Compression} shortens those instructions separately. At \emph{inference}, the Generator receives only the static prompt, question, and table definitions. Reference SQL is available to the offline judges and Refiner, never to the Generator. The judges assess SQL, the Refiner edits instructions, and the guard checks earlier successes before a revision is accepted.

\paragraph{Generate and judge.}
The Generator answers a training question using the current prompt and DDL. Three judges compare this SQL with the reference, given the same question and DDL. Each returns Yes, Maybe, or No with an explanation. Their fixed instructions cover semantics, timestamp types, windows, joins, and dialect validity. Result-set fields are left empty for the live-streaming evaluation protocol (\S\ref{sec:setup}). An example passes only when all three judges return Yes.

\paragraph{Rewrite from a failure.}
For a failing example, the Refiner receives the current prompt, fixed user template, question, DDL, reference and generated SQL, judge feedback, and summaries of failed attempts on that example. It returns a complete replacement prompt and a change summary. Instructions require construct-level rules without copying instance-specific identifiers or literals, and treat judge explanations as fallible evidence. Appendix~\ref{app:examples} examines a recorded edit. 

\paragraph{Check earlier successes.}
A candidate must first pass the target example. The guard then re-generates and re-judges examples fixed earlier in the current round and 12 randomly sampled passing examples. A candidate is accepted if at most one checked example fails. Rejected candidates leave the accepted prompt unchanged; their outcomes enter the attempt history. Up to five attempts are made per target. This sampled check controls acceptance while allowing limited regression.

\paragraph{Repeat and retain a checkpoint.}
Each round scores the induction sample, attempts repairs on up to 40 failing examples, and scores the resulting prompt again. The accepted prompt carries into the next round. After seven rounds, we retain round~6 using induction performance and saturation of prompt growth (\S\ref{sec:convergence}). Model assignments and sources are given in \S\ref{sec:setup}; Appendix~\ref{app:method} lists implementation settings.

% Queue the construct comparison for the top of page 4.
\begin{figure*}[t]
\centering
\includegraphics[width=\linewidth]{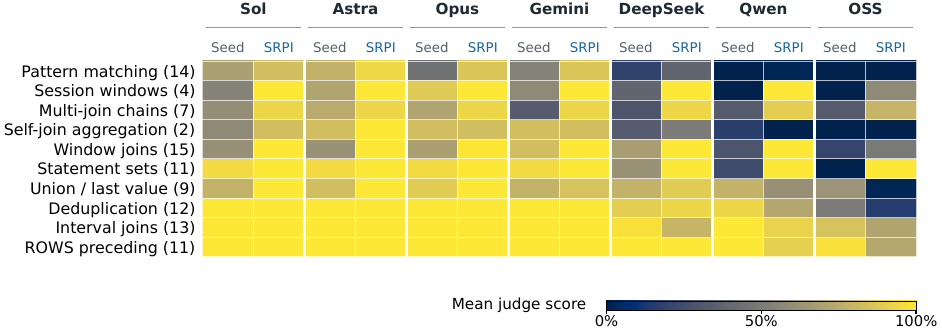}
\caption{Construct highlights: the four lowest seed means, four largest mean gains, and all four declines across seven Generators (10 unique labels). Adjacent cells show Seed/SRPI-60k scores (\%); parentheses give question counts. All 29 labels and values appear in Appendix~\ref{app:main}.}
\label{fig:miniheat}
\end{figure*}

\paragraph{Compress for inference.}
An early 50,000-character cap caused 123 rewrite rejections. We instead allow growth and compress afterwards. The Compressor receives only the prompt and a character budget, with instructions to preserve distinct constraints and rare-construct rules. It runs at temperature 0.1 for up to five attempts, aiming at 97\% of the budget. A candidate must be shorter and retain specified output-contract words. If the budget is missed, the shortest valid candidate is retained and its actual length reported. The 59,787-character version is denoted \textbf{SRPI-60k}.

\section{Evaluation}
\label{sec:exp}
We assess the system in Figure~\ref{fig:pipeline} for prompt transfer, induction and compression, and alternative feedback sources. Selected execution checks complement the judge scores.

\subsection{Protocol and metrics}
\label{sec:setup}
We compare the unchanged SRPI-60k prompt with the seed on seven Generators (Table~\ref{tab:main}), with one scored output per question and condition. GPT-6.1 Sol \citep{openai-2026-sol} performs induction, refinement, and compression. The other six test transfer: GPT-6 Astra \citep{openai-2026-astra}, Claude Opus 5.5 \citep{anthropic-2026-opus}, Gemini 3.8 Flash \citep{doshi-popa-2026-gemini}, DeepSeek 4.1 Flash \citep{deepseek-2026-flash}, Qwen3.8 27B \citep{qwen-2026-27b}, and GPT-OSS 120B \citep{openai-2025-gptoss}. Astra, Claude, and Gemini also form the judge panel. Appendix~\ref{app:method} details model attribution. For reporting, Yes, Maybe, and No map to 100\%, 50\%, and 0\%. \textbf{Micro} averages these scores over judges and questions. \textbf{Unanimous Yes} (U-Yes) is the percentage of questions receiving Yes from all three judges. Tier scores average judge scores within each tier; differences use percentage points (pp). Appendix~\ref{app:bar-results} retains construct-balanced Macro results. Judges accept semantically equivalent SQL formulations; scores are not calibrated correctness probabilities.

We use the configurable LLM judge in the Text-to-SQL Evaluation Toolkit \citep{hassanzadeh-etal-2026-toolkit}. Judges receive the question, SQL, and DDL without result sets. Live gold and generated queries can observe different records and temporal boundaries; arrival timing and state retention can also affect outputs \citep{confluent-2026-determinism}. Comparing these unaligned results would confound query quality with execution context. SQL-level judging provides a common assessment across the workload, complemented by cluster acceptance. The 95\% bootstrap intervals resample questions 10,000 times, conditional on recorded outputs.

\subsection{Transfer across Generators}
\label{sec:main}

SRPI improves every Generator's micro score by 7.1--19.2~pp, Tier~1 score by 3.7--17.3~pp, and Tier~3 score by 21.2--53.7~pp (Table~\ref{tab:main}). DeepSeek has the largest micro gain, rising from 70.0\% to 89.2\%. GPT-6.1 Sol improves from 85.3\% to 97.5\%, with a paired 95\% bootstrap interval of $[8.8,15.6]$~pp for the gain. Its unanimous-Yes rate rises from 72.0\% to 95.7\%. Final scores of 53.5--97.5\% still reflect differences in Generator capability. Appendix~\ref{app:bar-results} plots all metric and tier comparisons.

Four Generators exceed the provisional Tier~1 and Tier~3 thresholds on \emph{mean judge-score point estimates}. This statement changes under the stricter unanimity measure: Tier~1 unanimous-Yes rates are 95.8\%, 94.2\%, 94.8\%, and 94.2\% for Sol, Astra, Claude, and Gemini, respectively. Only Sol exceeds 95\% by that measure. Tier~3 contains 22 examples; unanimity offers a stricter development criterion.

Across all 29 labels, scores improve for 24, remain unchanged for one, and decline for four when averaged over Generators. From the rounded per-label scores, statement sets rise from 67\% to 100\%, and multi-join chains from 48\% to 91\%. Figure~\ref{fig:miniheat} highlights difficult constructs, large gains, and every declining label. Pattern matching remains difficult for several Generators. Appendix~\ref{app:main} contains the complete heatmap.

\subsection{Induction, convergence, and compression}
\label{sec:convergence}
\begin{figure*}[t]
\centering
\includegraphics[width=\linewidth]{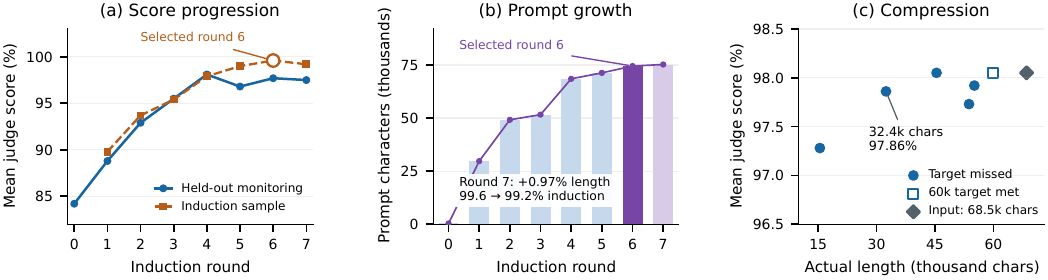}
\caption{Induction and compression (GPT-6.1 Sol). (a) Induction and domain-held-out scores; the latter track generalisation. (b) Prompt length by round. Round~6 is selected using induction performance and saturation of prompt growth; round~7 adds 0.97\% characters while induction performance falls. (c) A separate compression study starts from a 68,497-character prompt and reports actual lengths; only the 60k target is met. Score axes use different ranges.}
\label{fig:curve}
\label{fig:compress}
\end{figure*}
Figure~\ref{fig:curve} tracks seven rounds. We select round~6 using induction-set performance and saturation of prompt growth: its score reaches 99.6\%, then round~7 adds only 721 characters to 74,513 (0.97\%) while the score falls to 99.2\%. The held-out curve tracks generalisation and is not a checkpoint-selection criterion. This is a refinement trajectory, not a learning curve over training-set sizes.

\paragraph{Regression burden.}
In rounds 3--7, 88 candidate rewrites pass their target and 52 (59.1\%) are rejected by the guard. Target-local success therefore often co-occurs with sampled regressions, although stochastic generation and judging may contribute. A no-guard comparison would be needed to quantify its causal benefit.

\paragraph{Compression.}
In a separate compression study, a 68,497-character parent shrinks to 32,400 characters (52.7\% shorter), while micro changes from 98.05\% to 97.86\% in one scoring run. At 15,420 characters it is 97.28\%. Five of six requested budgets are missed; these single-run measurements describe an observed trade-off, not established non-inferiority. Transfer uses the 59,787-character SRPI-60k version. Appendix~\ref{app:prompts} provides fixed workflow prompts and selected induced-rule excerpts; Appendix~\ref{app:compress} gives all compression results.

\subsection{Alternative feedback sources}
\label{sec:ablation}
We compare prompts constructed from the same seed using 54 documentation pages, 18 parser-grammar slices, or three rounds of live execution errors on 300 induction examples. SRPI achieves the highest judge score among these configurations (Table~\ref{tab:ablation}). Lengths, supervision, and budgets differ; these are development alternatives, not controlled ablations or implementations of DIAL \citep{zhang-etal-2026-dial} or TAHOE \citep{chen-etal-2026-tahoe}.
\begin{table}[t]
\centering\small\setlength{\tabcolsep}{3.5pt}
\begin{tabular}{@{}lrrrr@{}}
\toprule
Prompt source & Chars & Micro\% & $\Delta$ pp & U-Yes\%\\
\midrule
Seed & 337 & 84.2 & -- & 71.2\\
Documentation & 19.9k & 85.2 & 1.0 & 74.3\\
Parser grammar & 18.5k & 84.9 & 0.7 & 73.5\\
Execution errors & 19.2k & 85.8 & 1.6 & 76.3\\
SRPI, uncompressed & 68.5k & \textbf{98.1} & \textbf{13.9} & \textbf{96.5}\\
\bottomrule
\end{tabular}
\caption{Alternative prompt sources, GPT-6.1 Sol. \textbf{Bold}: best score; $\Delta$ is in pp. SRPI uses reference SQL and more optimisation, so the comparison does not isolate the feedback source.}
\label{tab:ablation}
\end{table}
Reported cluster acceptance is 78.3\% for the seed, 85.2\% for execution-error rules, and 94.2\% for SRPI-60k. This indicates fewer rejected statements; acceptance alone does not establish result equivalence. Documentation and execution feedback may still complement SRPI.

\subsection{Panel sensitivity and execution checks}
We examine panel sensitivity and selected execution cases to understand remaining errors.

The evaluation comprises 1,799 outputs per prompt condition, each assessed by three judges. To probe panel sensitivity, we exclude a Generator's own judge where applicable: Astra improves from 89.8\% to 98.1\%, Claude from 83.9\% to 96.9\%, and Gemini from 78.7\% to 94.9\%. Gains therefore persist when these Generators do not score their own outputs. This check addresses self-evaluation within the existing panel, not validation by independent judges. Since induction and evaluation reuse the panel, optimisation toward its preferences remains possible; Sol also shares a model family with Astra.

Three selected seed outputs with split verdicts illustrate the distinction. Two \sql{TIMESTAMPDIFF} queries mix timestamp types and fail on the cluster, agreeing with the majority rejection. A \sql{SESSION} query also receives majority rejection but executes successfully. The cases help distinguish engine acceptance from semantic judgment. A schema-neutral rule diff also exposes reference bias (Appendix~\ref{app:examples}).

\section{Related Work}
\paragraph{Dialect adaptation.}
CrackSQL translates dialects, Dialect-SQL uses an intermediate representation, and SQL-GEN uses synthetic data and model merging \citep{zhou-etal-2025-cracksql,shi-etal-2025-dialect,pourreza-etal-2024-sqlgen}. Query-plan approaches compile dialect-agnostic relational plans into SQL \citep{royer-etal-2026-plans}, requiring compiler support for each backend; SRPI retains direct SQL generation and adds prompt context. DIAL combines dialect-aware planning, a hierarchical knowledge base, execution repair, and semantic verification \citep{zhang-etal-2026-dial}. TAHOE retrieves hints learned from compiler and semantic feedback \citep{chen-etal-2026-tahoe}. SRPI instead consolidates guidance offline, requiring neither retrieval nor repair at inference.

\paragraph{Prompt optimisation.}
APE and OPRO select prompts by score \citep{zhou-etal-2023-ape,yang-etal-2024-opro}; ProTeGi combines textual gradients, beam search, and bandit selection \citep{pryzant-etal-2023-automatic}; TextGrad propagates feedback through computation graphs \citep{yuksekgonul-etal-2025-textgrad}; GEPA reflects on trajectories and evolves a Pareto frontier \citep{agrawal-etal-2025-gepa}. ACE incrementally updates a playbook to avoid context collapse \citep{zhang-etal-2025-ace}. SRPI retains one accepted prompt, rewrites it per failure using three judges' explanations and attempt history, checks regressions, and compresses separately. Self-Refine revises outputs, Reflexion maintains agent memory, and STOP modifies the optimisation scaffold \citep{madaan-etal-2023-selfrefine,shinn-etal-2023-reflexion,zelikman-etal-2023-stop}. SRPI's judges, Refiner instructions, and guard stay fixed: optimisation is single-level and externally supervised.

\paragraph{SQL judging.}
LLM judges and panels provide scalable feedback \citep{zheng-etal-2023-judge,verga-etal-2024-poll}. Distilled test suites broaden execution coverage \citep{zhong-etal-2020-semantic}; FLEX uses execution-informed judges \citep{kim-etal-2025-flex}. The Text-to-SQL Evaluation Toolkit combines execution and LLM metrics \citep{hassanzadeh-etal-2026-toolkit}; we use SQL-level judging. Judge-specific overfitting during prompt optimisation has been observed in legal QA \citep{elganayni-etal-2026-disposition}, motivating separate scrutiny of evaluator transfer.

\section{Discussion and Conclusion}
SRPI converts generation failures into explicit dialect specifications. Its 24 prompt sections cover timestamp compatibility, windows, joins, NULL handling, numeric contracts, and a final checklist. Gains across seven Generators and held-out domains support reuse of this guidance beyond the induction examples. The resulting artifact can be versioned or rolled back without retraining models or adding inference stages. The transferable unit is a SQL constraint. The Refiner expresses such constraints through schema-neutral placeholders for reuse by the Generator across domains. Curated failures can guide later offline prompt updates.

\paragraph{Monitor construct-level behaviour.}
Session-window TVFs improve from 44\% to 94\% averaged across Generators, whereas deduplication and interval joins decline by about 8 and 6~pp. The guard rejects 52 of 88 target-passing rewrites, showing why a local repair merits checks on earlier successes. These observations support retaining representative passing cases alongside new failures. Versioning prompt, Generator, and engine together would make later changes traceable.

\paragraph{Budget the whole request.}
SRPI-60k adds about 7,400 input words per call. Mean completion tokens fall from 248.1 to 209.6 for Sol ($-15.5\%$), but rise from 1,131.2 to 2,915.4 for DeepSeek ($+157.7\%$) and 1,415.3 to 3,187.2 for Gemini ($+125.2\%$). Model choice therefore affects output size as well as quality. Compression measures a prompt-character/score trade-off; input-token costs, billing, and latency remain unmeasured (Appendix~\ref{app:main}).

\paragraph{Early access and continued refinement.}
The assistant using the induced prompt is in early access. Encouraging informal feedback from Flink and streaming-data experts motivates further evaluation; initial real-user feedback is forthcoming. We plan to curate recurring failures from user query logs into question--DDL examples with validated reference SQL, refine the prompt offline, and regression-check versioned replacements before release. This would extend the evaluated workflow into an ongoing maintenance process. SRPI serves the generation component after schema selection; the surrounding assistant can retain schema retrieval and resource controls. The frozen prompt requires one Generator call without runtime retrieval or repair. Future user-log refinements will be evaluated separately from the controlled results reported here. User feedback will guide priorities for expanding construct coverage.

\label{main-text-end}

\clearpage
\section*{Limitations}
\textbf{Evaluation and deployment stage.} SQL-level judge scores and cluster acceptance provide complementary development evidence. Measurements are offline; real-user feedback and operational outcomes are forthcoming. Expert feedback is informal, without a documented human evaluation. Induction and evaluation share judges in a SQL-only configuration; systematic human scoring and quantified human--judge disagreement are absent. Independent scoring or a disjoint panel could clarify semantic errors. Live result-set comparisons require aligned inputs and temporal boundaries; controlled replay or fixed-input tests remain possible complements.

\textbf{Workload coverage and intended generalisation.} The controlled workload tests reuse across domains and Generators within a defined construct repertoire. Domains and exact SQL instances are disjoint, but pattern templates overlap. This tests reuse of failure-derived rules when recurring windowing and join patterns are applied across schemas. Unseen-template generalisation and representativeness of production traffic are untested; mishandled constructs may occur in pretraining. Explicit schema names, three held-out domains and 22 Tier~3 questions delimit the evidence. Schema ambiguity, multi-turn interaction and other engines remain untested.

\textbf{Experimental variability.} Results reflect one induction run and one scored output per question and condition. Bootstrap intervals measure question variability, not model sampling or future versions. Guard activity is reported without isolating rewrite, panel or guard effects; compression and cluster acceptance are single-run measurements. Matched-budget comparisons with SFT, other optimisers and retrieval/repair pipelines, plus training-set-size ablations, remain open.

\textbf{Operational trade-offs and artifacts.} Single-call inference simplifies serving but adds prompt context; end-to-end cost and latency are unmeasured. The sampled guard permits limited regressions, and anti-copying instructions cannot eliminate reference-derived assumptions. The supplement supplies fixed workflow prompts, selected learned-rule excerpts, a recorded diff and request, numerical tables and plotting code (Appendix~\ref{app:prompts}). Full induced/compressed prompts, induction logs and per-example evaluation records are absent, so artifacts support inspection rather than exact reruns.

\newpage
\section*{Ethical Considerations}
The study uses synthetic data rather than customer or personal records. Generated SQL may still expose information or trigger costly or destructive operations when deployed; the workload includes sink and background statements as well as reads. Appropriate permissions, explicit approval of writes, resource limits, and reviewable SQL are therefore necessary deployment safeguards, not capabilities established by this study. Planned user-log curation will require access controls and removal of customer-sensitive content before examples are sent to hosted models. The multi-model offline loop incurs unmeasured computational and energy costs.

\bibliography{custom}
\appendix
\raggedbottom
% Supplementary details. All numeric results are retained from the supplied study.

\paragraph{Artifact inventory.}
The supplement includes fixed workflow prompts, the seed and user template, three learned-rule excerpts, a recorded diff and Refiner request, numerical tables, and plotting code. Complete induced/compressed prompts, full induction logs, and per-example evaluation records are not included. \path{ARTIFACT_MANIFEST.txt} maps supplied files to the analyses. Appendix~\ref{app:prompts} reproduces the fixed prompts; Appendix~\ref{app:compression-examples} labels the compaction examples as editorial illustrations, not recorded Compressor outputs.

\section{Data construction and split}
\label{app:data}

\subsection{Synthetic schemas and stream data}
The seed of the pipeline is a set of 30 question--SQL pairs written by experts and run end to end on a live Confluent Cloud Flink cluster.
The schema portfolio has 50 schemas over 45 business domains, with 304 tables and 1,752 columns. Schemas are drawn from ten structural families
(normalised entity networks, header--detail, many-to-many, hierarchies and graphs, supertype--subtype, star, snowflake, fact constellations,
temporal/event history, semi-structured data) and sixteen features (composite keys, alternative join paths, temporal histories, nested data, and others).
For every portfolio slot, an LLM receives the specification format, summaries of the other slots and a list of Flink capability constraints, and
writes the grain, columns, keys, relations and value rules of each table, plus a real-time component. A deterministic validator checks keys, types, tenant-scoped joins
and watermark columns, and an LLM structure judge checks that the intended family and features are really exercised. Accepted specifications are compiled
deterministically to Flink DDL (primary keys \sql{NOT ENFORCED}, event-time watermarks).

Tables are populated by workflows with three strategies: \emph{seed} (fixed rows for small reference tables), \emph{Faker} (bounded, append-only Flink Faker sources followed by an
\sql{INSERT ... SELECT} that derives dependent columns) and \emph{derived} (child records and events computed from populated tables in snapshot mode).
A linter checks the workflows. Live SQL assertions test row counts, key uniqueness, foreign keys, allowed values, ranges and domain rules. A further LLM review
checks the plausibility of values and times. In the documented 50-schema run all 4,207 live integrity checks passed and the realism loop reduced error-level findings
from 52 to 4.

\subsection{Pattern taxonomy}
The 42 SQL patterns come from the 30 human pairs and from the official documentation and how-to guides, and are organised in 14 groups (Table~\ref{tab:groups}).
The construct labels used for macro scores and heatmaps are the pattern names; 31 of them occur in the pool and 29 in the evaluation set.

\begin{table*}[t]
\centering\footnotesize\setlength{\tabcolsep}{3pt}
\begin{tabular}{@{}l >{\raggedright\arraybackslash}p{0.7\textwidth}@{}}
\toprule
Group & Patterns \\
\midrule
Windowed aggregation & \sql{tumble\_tvf}, \sql{hop\_tvf}, \sql{session\_window\_tvf}, \sql{windowed\_count\_having} \\
Window join & \sql{window\_join} \\
OVER windows & \sql{over\_rows\_preceding}, \sql{over\_range\_interval\_preceding}, \sql{over\_window\_unbounded}, \sql{lag\_over\_window}, \sql{lead\_over\_window}, \sql{avg\_over\_partition} \\
Dedup / Top-N & \sql{dedup\_rownumber\_eq1}, \sql{streaming\_topn\_rownumber} \\
Stream--stream joins & \sql{stream\_stream\_interval\_join}, \sql{unbounded\_inner\_join}, \sql{unbounded\_left\_join} \\
Temporal / enrichment & \sql{temporal\_join}, \sql{multi\_join\_chain}, \sql{two\_table\_join\_agg}, \sql{self\_join\_aggregation} \\
Filter / data quality & \sql{filter\_with\_interval\_bound}, \sql{null\_empty\_filter} \\
Snapshot queries & \sql{group\_by\_order\_limit}, \sql{set\_snapshot\_mode} \\
CEP & \sql{match\_recognize}, \sql{union\_all\_last\_value} \\
JSON / serialisation & \sql{json\_value\_extraction}, \sql{json\_value\_cast}, \sql{metadata\_virtual\_column} \\
UDFs / extensibility & \sql{scalar\_udf\_call}, \sql{table\_udf\_lateral}, \sql{ptf\_call}, \sql{ml\_over\_window} \\
Sink / background & \sql{execute\_statement\_set}, \sql{ctas\_passthrough}, \sql{regexp\_replace\_mask} \\
Schema definition & \sql{set\_state\_ttl}, \sql{faker\_connector\_ddl}, \sql{create\_function\_jar} \\
Scalar expressions & \sql{convert\_tz}, \sql{timestampdiff}, \sql{time\_series\_math} \\
\bottomrule
\end{tabular}
\caption{The 14 pattern groups and their 42 patterns.}
\label{tab:groups}
\end{table*}

\subsection{Cluster-grounded generation and splits}
Given the populated environment, the pipeline lists the Kafka topics and reads their schemas from Schema Registry, writes Flink DDL with watermark and metadata columns, and instantiates the pattern templates
against the schemas. Every candidate statement is submitted to the cluster, and failures are discarded. An LLM then writes the natural-language question at several levels of explicitness;
the experiments use the form that explicitly names tables and columns. This reduces schema-linking ambiguity relative to unconstrained user requests. Labels and tiers are assigned automatically from the pattern(s) a query uses.
Gold queries were run on the cluster when the pool was built. Their result sets exist for 73\% of the pool and are capped at 200 rows. They are stored with the dataset but are not shown to the judges. These outputs were collected during pool construction, not as paired executions with each generated query. Even snapshot queries read the current state, so separate executions can use different input cutoffs \citep{confluent-2026-snapshot}. Streaming outputs can additionally depend on arrival timing and retained state \citep{confluent-2026-determinism}. Supplying such unaligned, truncated results to a judge would mix query differences with execution-context differences. We therefore configure the toolkit to assess the SQL with the question and DDL; controlled replay or fixed-input testing would be a separate evaluation protocol.

\begin{table*}[t]
\centering\small
\begin{tabular}{@{}lrrr@{}}
\toprule
 & Pool & Train & Eval. \\
\midrule
Questions & 3,940 & 3,683 & 257 \\
Domains & 45 & 42 & 3 \\
Labels & 31 & 31 & 29 \\
Tier 1 / 2 / 3 & 2,864 / 639 / 437 & 2,673 / 595 / 415 & 191 / 44 / 22 \\
Snapshot / streaming / background & 2,432 / 1,356 / 152 & -- & 154 / 92 / 11 \\
\bottomrule
\end{tabular}
\caption{Pool, training pool and evaluation set. Evaluation domains: education (67), insurance (98), retail (92). The evaluation mean question length is about 45 words and the mean gold SQL
length about 580 characters. Tier 3 in the evaluation set has one 5-label, eleven 4-label and ten 3-label queries.}
\label{tab:splits}
\end{table*}

Domains and exact SQL strings are disjoint between the training pool and evaluation set, while pattern templates are shared. The controlled split tests whether guidance induced from failures on known construct families transfers to new domain instances. Evaluation across Generators separately tests reuse of the same frozen instructions by other LLMs. Unseen-template generalisation is outside this experimental objective. Checkpoint selection uses induction-set performance and saturation of prompt growth (Appendix~\ref{app:loop}); the evaluation curve is used only to track generalisation.

The induction sample of 335 questions has 222 / 71 / 42 questions per tier, 42 domains and 31 labels. It starts from 300 questions stratified by label and domain; 35 further questions
are added for 13 long-tail labels. Round~1 of the run used the 300 base questions and the sample was extended to 335 from round~2.

\section{Implementation details}
\label{app:method}

\subsection{Models and source attribution}
Table~\ref{tab:model-sources} identifies the model releases and their roles. All seven generate SQL in the transfer evaluation. Sol also performs induction, refinement, and compression; Astra, Claude, and Gemini form the judge panel. Release citations identify the named models; exact serving snapshots and run dates are not available in the supplied records.

\begin{table*}[t]
\centering\small\setlength{\tabcolsep}{5pt}
\begin{tabular}{@{}p{0.23\textwidth}p{0.40\textwidth}p{0.30\textwidth}@{}}
\toprule
Model & Role & Release or model-card source \\
\midrule
GPT-6.1 Sol & Induction, refinement, compression, evaluation & \citet{openai-2026-sol} \\
GPT-6 Astra & Judge; transfer evaluation & \citet{openai-2026-astra} \\
Claude Opus 5.5 & Judge; transfer evaluation & \citet{anthropic-2026-opus} \\
Gemini 3.8 Flash & Judge; transfer evaluation & \citet{doshi-popa-2026-gemini} \\
DeepSeek 4.1 Flash & Transfer evaluation & \citet{deepseek-2026-flash} \\
Qwen3.8 27B & Transfer evaluation & \citet{qwen-2026-27b} \\
GPT-OSS 120B & Transfer evaluation & \citet{openai-2025-gptoss} \\
\bottomrule
\end{tabular}
\caption{Model attribution and roles. Sources refer to public releases, not inferred experimental snapshot identifiers.}
\label{tab:model-sources}
\end{table*}

\subsection{Hyperparameters}
\begin{table}[H]
\centering\small
\begin{tabular}{@{}lr@{}}
\toprule
Generator temperature & 0.2 \\
Refiner temperature / max output tokens & 0.4 / 24,000 \\
Refiner format retries & 2 \\
Attempts per failing instance & 5 \\
Failing instances repaired per round & $\le 40$ \\
Guard sample (random passing instances) & 12 \\
Guard tolerance (flips allowed) & 1 \\
Compressor temperature / target / attempts & 0.1 / 0.97$L$ / 5 \\
\bottomrule
\end{tabular}
\caption{Settings of the run. A Refiner reply without a \texttt{<system\_prompt>} block, or a prompt identical to its input, is rejected.}
\end{table}

An instance with only infrastructure errors (a failed generation or judge call) is reported to the Refiner as ``no verdict; retry'' and is not used as evidence about the prompt.
The Compressor accepts a candidate only if it is shorter than its input and still contains the output-contract words (\emph{fenced}, \emph{code block}, \emph{only}, \emph{one}).
If the limit is missed after the last attempt, the shortest valid prompt is kept and scored anyway.

A candidate that fails the target may serve as the next local rewrite base when its verdicts are informative, but it is not committed as the accepted prompt.

Appendix~\ref{app:prompts} reproduces the complete static workflow prompts. Appendix~\ref{app:rule-excerpts} gives induced-rule excerpts and Appendix~\ref{app:compression-examples} illustrates the compression instructions.

\section{Induction trajectory}
\label{app:loop}

The number of instances attempted falls from 31 in round~2 to 2 in round~7, and the Refiner needs fewer attempts: 23 of the 36 instances fixed in rounds 3--7 were fixed by the first rewrite.
The share of rewrites rejected by the guard is highest in rounds 3 and 4 (36 of 46 and 13 of 28 passing candidates). Rounds 1--2 and 3--7 come from two consecutive stretches of one run,
and round~3 starts from the round-2 prompt. We select round~6 from induction performance and saturation of prompt growth, without using the evaluation curve for selection. The induction sample changes from 300 to 335 questions after round~1. From round~6 to 7, prompt length increases by only 721 characters (0.97\%) while induction micro decreases from 99.6\% to 99.2\%. These observations indicate diminishing returns and support retaining round~6. The compression study below independently reports measurements for its 68,497-character input prompt.

\begin{table*}[t]
\centering\small\setlength{\tabcolsep}{3.5pt}
\begin{tabular}{@{}rrcccrr@{}}
\toprule
Round & Chars & \multicolumn{2}{c}{Eval. micro / macro (\%)} & Full agr.\ (\%) & Ind.\ micro (\%) & Ind.\ $n$ \\
\midrule
0 (seed) & 337    & 84.2 & 84.8 & 71.2 & -- & -- \\
1 & 29,701 & 88.8 & 89.5 & 79.0 & 89.8 & 300 \\
2 & 49,179 & 92.9 & 94.1 & 86.4 & 93.7 & 335 \\
3 & 51,614 & 95.5 & 97.1 & 92.6 & 95.4 & 335 \\
4 & 68,497 & \textbf{98.1} & \textbf{98.8} & \textbf{96.5} & 97.9 & 335 \\
5 & 71,294 & 96.8 & 97.9 & 92.2 & 99.0 & 335 \\
6 & 74,513 & 97.7 & 98.6 & 95.7 & \textbf{99.6} & 335 \\
7 & 75,234 & 97.5 & 98.5 & 95.3 & 99.2 & 335 \\
\bottomrule
\end{tabular}
\caption{Evaluation and induction-sample scores after each round. \emph{Full agr.}: share of answers where all three judges say Yes. Bold marks the largest value in each score column, not a checkpoint-selection criterion.}
\label{tab:curve}
\end{table*}

\begin{table*}[t]
\centering\small\setlength{\tabcolsep}{3.5pt}
\begin{tabular}{@{}rrrrrrrr@{}}
\toprule
Round & Chars & Failing & Fixed & Refiner calls & Passed target & Guard-rej. & Lines kept (\%) \\
\midrule
% Final column is percentage of lines kept; other numeric columns are counts.
1 & 29{,}701 & 20 & 4 & -- & -- & -- & 0 \\
2 & 49{,}179 & 31 & 8 & -- & -- & -- & 11 \\
3 & 51{,}614 & 18 & 10 & 57 & 46 & 36 & 6 \\
4 & 68{,}497 & 17 & 15 & 38 & 28 & 13 & 12 \\
5 & 71{,}294 & 8 & 6 & 17 & 7 & 1 & 59 \\
6 & 74{,}513 & 5 & 4 & 13 & 6 & 2 & 80 \\
7 & 75{,}234 & 2 & 1 & 6 & 1 & 0 & 97 \\
\bottomrule

\end{tabular}
\caption{Repair activity per round. \emph{Failing}: instances attempted; \emph{Fixed}: instances for which a rewrite was accepted;
\emph{Passed target}: Refiner calls whose output passed the target instance; \emph{Guard-rej.}: of these, rejected by the guard;
\emph{Lines kept}: share of the non-empty lines of the new prompt that occur verbatim in the previous prompt.
For rounds 1--2, the calls rejected for exceeding the old 50,000-character limit were not logged individually (123 rejections in that earlier setting), so the call counts are omitted.}
\label{tab:loopstats}
\end{table*}

\section{Compression results}
\label{app:compress}

\begin{table*}[t]
\centering\small\setlength{\tabcolsep}{4pt}
\begin{tabular}{@{}rrcccc@{}}
\toprule
Limit & Reached & Micro (\%) & Macro (\%) & Full agr.\ (\%) & Limit met \\
\midrule
(parent) & 68,497 & \textbf{98.05} & \textbf{98.80} & 96.1 & -- \\
60k & 59,787 & \textbf{98.05} & 98.75 & \textbf{96.5} & yes \\
50k & 55,088 & 97.92 & 98.27 & 95.7 & no \\
40k & 53,717 & 97.73 & 98.79 & 95.3 & no \\
30k & 45,372 & \textbf{98.05} & \textbf{98.80} & \textbf{96.5} & no \\
20k & 32,400 & 97.86 & 98.36 & 96.1 & no \\
10k & 15,420 & 97.28 & 96.63 & 94.2 & no \\
\midrule
Seed prompt & 337 & 84.7 & 84.9 & 71.6 & -- \\
\bottomrule
\end{tabular}
\caption{A separate compression study with a 68,497-character input prompt, GPT-6.1 Sol, and 257 evaluation questions, one scoring per prompt. The parent scored 98.05\%; the seed prompt was scored again in this experiment (84.7\%) and differs from its other scoring (84.2\%). Bold marks the largest score in each column; ties are retained.}
\label{tab:compress}
\end{table*}

The Compressor reaches only the 60k limit. The first pass lands at about 61k, 59k, 59k, 53k, 41k and 21k characters for the six limits, and later passes gain little.
Within the flat region, differences of 0.3 percentage points or less are smaller than the re-scoring differences observed elsewhere in the study. Each compression candidate was scored once; there is no paired non-inferiority analysis or repeated-run estimate.

\section{Additional evaluation results}
\label{app:main}

\subsection{Tier scores for all Generators}
\begin{table*}[t]
\centering\small\setlength{\tabcolsep}{3pt}
% Percent-scale metrics. Bold marks the largest reported value in each metric column; ties are retained.
\begin{tabular}{@{}lrrrrrr@{}}
\toprule
 & \multicolumn{2}{c}{Tier 1 (\%; $n{=}191$)} & \multicolumn{2}{c}{Tier 2 (\%; $n{=}44$)} & \multicolumn{2}{c}{Tier 3 (\%; $n{=}22$)} \\
\cmidrule(lr){2-3}\cmidrule(lr){4-5}\cmidrule(lr){6-7}
Generator & Seed & SRPI & Seed & SRPI & Seed & SRPI \\
\midrule
GPT-6.1 Sol & 87.2 & 97.3 & \textbf{85.6} & 97.0 & 68.9 & \textbf{100.0} \\
GPT-6 Astra & 88.3 & 96.9 & 85.2 & \textbf{98.1} & \textbf{69.7} & \textbf{100.0} \\
Claude Opus 5.5 & \textbf{88.8} & \textbf{97.5} & 77.7 & 96.6 & \textbf{69.7} & 94.7 \\
Gemini 3.8 Flash & 85.3 & 96.5 & 75.8 & 95.5 & \textbf{69.7} & 90.9 \\
DeepSeek 4.1 Flash & 72.6 & 89.9 & 62.1 & 87.1 & 62.9 & 87.9 \\
Qwen3.8 27B & 66.1 & 77.3 & 55.7 & 75.0 & 31.1 & 84.8 \\
GPT-OSS 120B & 49.7 & 53.4 & 48.1 & 61.0 & 13.6 & 39.4 \\
\bottomrule
\end{tabular}

\caption{Mean judge score (\%) by tier, seed prompt vs.\ SRPI-60k. Bold marks the largest value in each column.}
\label{tab:tiers}
\end{table*}
\begin{table*}[t]
\centering\small\setlength{\tabcolsep}{3pt}
% Percent-scale metrics. Bold marks the largest reported value in each metric column; ties are retained.
\begin{tabular}{@{}lrrrrrr@{}}
\toprule
 & \multicolumn{2}{c}{Tier 1 (\%)} & \multicolumn{2}{c}{Tier 2 (\%)} & \multicolumn{2}{c}{Tier 3 (\%)} \\
\cmidrule(lr){2-3}\cmidrule(lr){4-5}\cmidrule(lr){6-7}
Generator & Seed & SRPI & Seed & SRPI & Seed & SRPI \\
\midrule
GPT-6.1 Sol & 73.8 & \textbf{95.8} & 72.7 & 93.2 & 54.5 & \textbf{100.0} \\
GPT-6 Astra & \textbf{77.5} & 94.2 & \textbf{75.0} & \textbf{95.5} & 54.5 & \textbf{100.0} \\
Claude Opus 5.5 & 76.4 & 94.8 & 63.6 & 90.9 & 54.5 & 90.9 \\
Gemini 3.8 Flash & 73.8 & 94.2 & 63.6 & 88.6 & \textbf{63.6} & 81.8 \\
DeepSeek 4.1 Flash & 59.7 & 82.7 & 50.0 & 79.5 & 50.0 & 77.3 \\
Qwen3.8 27B & 52.9 & 72.3 & 45.5 & 72.7 & 13.6 & 77.3 \\
GPT-OSS 120B & 37.7 & 45.0 & 38.6 & 52.3 & 13.6 & 36.4 \\
\bottomrule
\end{tabular}

\caption{Full agreement (\% of answers) by tier. Tier sizes are 191, 44 and 22. Bold marks the largest value in each column.}
\label{tab:cata-tiers}
\end{table*}

\paragraph{Confidence intervals.}
95\% bootstrap intervals (10,000 resamples of the 257 questions, GPT-6.1 Sol, GPT-6 Astra, Claude Opus 5.5, Gemini 3.8 Flash): micro with SRPI-60k (\%) [95.8, 98.9], [95.8, 98.8], [95.5, 98.5] and [93.8, 97.7];
paired gain over the seed prompt (percentage points) [8.8, 15.6], [8.0, 14.7], [8.7, 15.0] and [9.9, 17.4]. Tier~1 (\%) [95.2, 99.1], [94.9, 98.7], [95.7, 99.0] and [94.1, 98.5].
Tier~3 (\%; 22 questions) [100, 100], [100, 100], [84.8, 100] and [81.8, 98.5]. These intervals resample questions from one run; they do not quantify judge error or run-to-run variation. The degenerate Tier~3 intervals arise because all observed scores equal 100\%; they do not establish perfect population performance.

\subsection{Judge-panel sensitivity}

\begin{table}[H]
\centering\footnotesize\setlength{\tabcolsep}{4pt}
\fitcol{\begin{tabular}{@{}lcccc@{}}
\toprule
Judge scoring its own outputs & seed own & seed others & SRPI own & SRPI others \\
\midrule
GPT-6 Astra & 79.0 & \textbf{89.8} & 96.1 & \textbf{98.1} \\
Claude Opus 5.5 & 87.9 & 83.9 & 97.5 & 96.9 \\
Gemini 3.8 Flash & \textbf{89.5} & 78.7 & \textbf{97.7} & 94.9 \\
\bottomrule
\end{tabular}}
\caption{Mean judge score (\%) of a model's own outputs, assessed by itself and by the other two judges. Bold marks the largest score per column; it is not a measure of judge validity.}
\end{table}

Excluding the Generator itself from the panel gives SRPI-60k micro scores of 98.1\%, 96.9\% and 94.9\% for GPT-6 Astra, Claude Opus 5.5 and Gemini 3.8 Flash, respectively (seed: 89.8\%, 83.9\%, 78.7\%). GPT-6.1 Sol is not a judge: its reported 97.5\% (seed: 85.3\%) uses the complete panel and is not a leave-one-family-out result; GPT-6 Astra remains in its panel. These checks measure sensitivity to panel membership, not agreement with independently established correctness.

\paragraph{Labels with lower scores.}
Averaged over the seven Generators, 24 of the 29 labels improve, one is unchanged, and four score lower with SRPI-60k: \sql{dedup\_rownumber\_eq1} ($-7.6$ pp), \sql{stream\_stream\_interval\_join} ($-5.7$ pp), \sql{over\_rows\_preceding} ($-4.7$ pp) and \sql{union\_all\_last\_value} ($-1.4$ pp). These averages are calculated from the rounded per-model cells in Figure~\ref{fig:heat}; \sql{metadata\_virtual\_column} remains at 100\%.
Averaged over the seven Generators, \sql{session\_window\_tvf} goes from 44\% to 94\%.

\paragraph{Judge disagreement.}
Table~\ref{tab:judge-ex} juxtaposes judge verdicts and cluster outcomes for three selected evaluation answers of GPT-6.1 Sol with the seed prompt.
Instances 1006299 and 3706282 omit \sql{CAST(\$rowtime AS TIMESTAMP(3))} inside \sql{TIMESTAMPDIFF}, mixing \sql{TIMESTAMP} and \sql{TIMESTAMP\_LTZ}. Instance 2006212 uses \sql{FROM SESSION(...)} without an enclosing \sql{TABLE(...)}. The table records each judge's verdict.
The seed-prompt statements were also run on the cluster. Instances 1006299 and 3706282 fail there with \emph{TIMESTAMP\_LTZ only supports diff between the same type}, so the two judges who said No on 3706282 were right and the Yes on 1006299 was wrong. Instance 2006212 runs and returns rows, so the cluster accepts the \sql{SESSION} call without \sql{TABLE(...)} and the two No verdicts there are not confirmed by execution. The induced prompt contains a rule for both forms and its outputs pass all three judges.
\begin{table}[H]
\centering\footnotesize\setlength{\tabcolsep}{3pt}
\begin{tabular}{@{}lcccc@{}}
\toprule
Instance & Claude & Astra & Gemini & Engine \\
\midrule
1006299 & No & No & Yes & Fails \\
2006212 & No & Yes & No & Runs \\
3706282 & No & Yes & No & Fails \\
\bottomrule
\end{tabular}
\caption{Selected judge-disagreement cases, not a population estimate. The two timestamp queries fail on the cluster; the SESSION query runs and returns rows. Engine acceptance alone does not establish semantic equivalence.}
\label{tab:judge-ex}
\end{table}
These cases illustrate two corroborated typing errors and one unsupported syntax objection. They were selected for disagreement and contain no unanimous verdicts. Representative joint counts of judge verdicts and execution outcomes are not available, so they cannot estimate how often consensus conflicts with engine behaviour. Per-tier human--judge disagreement rates are not reported.

\subsection{Completion tokens}
\begin{table}[H]
\centering\footnotesize\setlength{\tabcolsep}{3pt}
\begin{tabular}{@{}lrrr@{}}
\toprule
Generator & Seed & SRPI-60k & Change (\%) \\
\midrule
Qwen3.8 27B & \textbf{190.8} & 195.9 & $+2.7$ \\
GPT-6 Astra & 239.7 & \textbf{201.0} & $-16.1$ \\
GPT-6.1 Sol & 248.1 & \textbf{209.6} & $-15.5$ \\
Claude Opus 5.5 & 503.3 & \textbf{493.8} & $-1.9$ \\
GPT-OSS 120B & \textbf{704.0} & 829.1 & $+17.8$ \\
DeepSeek 4.1 Flash & \textbf{1131.2} & 2915.4 & $+157.7$ \\
Gemini 3.8 Flash & \textbf{1415.3} & 3187.2 & $+125.2$ \\
\bottomrule
\end{tabular}
\caption{Average completion tokens per question; bold marks the lower count per model. Change is $100(\mathrm{SRPI}/\mathrm{Seed}-1)$, derived from the recorded averages, not repeated runs. Counts from different tokenizers need not be comparable.}
\label{tab:completion-tokens}
\end{table}
SRPI-60k adds about 7,400 words of prompt text to every request. Input-token counts, timings, billing records, and completion measurements across compression variants are not available. The table describes output-length changes; it does not establish an end-to-end cost or latency frontier.

\subsection{Per-label scores}

\begin{figure*}[p]
\centering
\includegraphics[width=\textwidth]{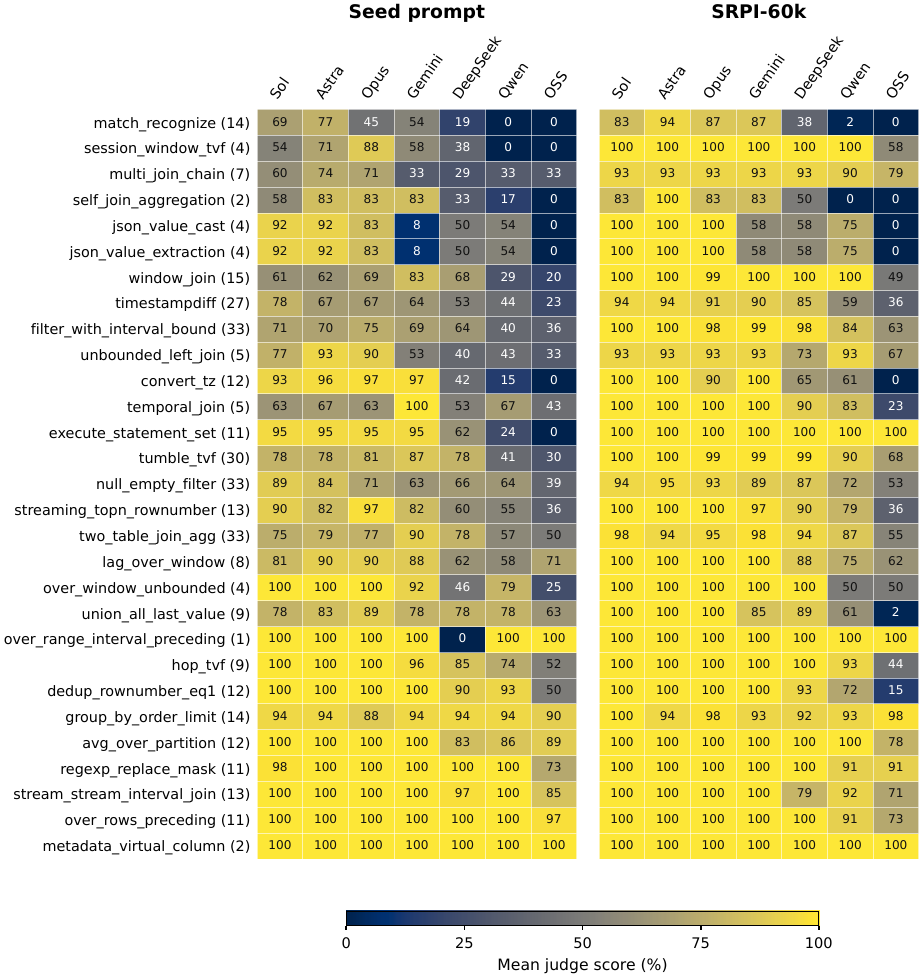}
\caption{Per-label mean judge scores for all seven Generators, seed (left) and SRPI-60k (right). Labels are sorted by seed-prompt mean; parentheses give the number of evaluation questions carrying each label. Scores are percentages, based on source values originally rounded to two decimal places on the unit scale. Short model names follow Table~\ref{tab:main}.}
\label{fig:heat}
\end{figure*}
Figure~\ref{fig:heat} displays the reported per-label values. The exact source tables are preserved in \path{generated/labels_v1.tex} and \path{generated/labels_rsi.tex}. Small supports, such as two questions for \sql{self\_join\_aggregation}, make individual label estimates unstable.

\section{Alternative feedback configurations}
\label{app:ablation}

\begin{table}[H]
\centering\footnotesize\setlength{\tabcolsep}{3pt}
\begin{tabular}{@{}p{0.27\columnwidth}p{0.62\columnwidth}@{}}
\toprule
Prompt & How it was built \\
\midrule
Documentation (19.9k) & An LLM agent read 54 pages of the official documentation, extracted 623 declarative rules and consolidated them to 164. \\
Parser grammar (18.5k) & An agent read 18 slices of the Calcite/Flink parser grammar, extracted 139 rules and consolidated them to 96. \\
Execution errors (19.2k) & The Generator's SQL was run on the live cluster for three rounds over 300 induction questions; each error class was turned into a general rule (16 rules). \\
SRPI (68.5k) & An uncompressed prompt produced from the seed by the loop of \S\ref{sec:method}. \\
\bottomrule
\end{tabular}
\caption{How the prompts of Table~\ref{tab:ablation} were produced. All start from the seed prompt; none include rules from another column.}
\end{table}

The share of evaluation queries that the cluster accepts (``execution rate'') was 78.3\% for the seed prompt and 85.2\% with the execution-error prompt; the execution-rate table of the study reports 94.2\% for SRPI-60k.
These rates come from the study's summary table. Execution acceptance is a weak proxy for correctness: a statement can run on the cluster and answer the wrong question,
so these rates cannot establish semantic correctness or explain the score difference by themselves.

Limits of this comparison: one run per prompt; prompt sizes differ (18.5k to 68.5k); SRPI had more optimisation steps and used the gold SQL, whereas the documentation and grammar prompts did not; and the three signals were not stacked.
The reported configurations favour SRPI on the judge metric. They do not isolate the causal contribution of the feedback source, compare with full DIAL \citep{zhang-etal-2026-dial} or TAHOE \citep{chen-etal-2026-tahoe} systems, or show that documentation and execution feedback are unnecessary.

\section{Induced rules and failure analysis}
\label{app:examples}

SRPI encodes failure-derived guidance in the prompt; the examples below illustrate both useful rules and reference bias. SRPI-60k has 59,787 characters and 24 sections. Its largest sections cover measures, NULLs and numeric contracts (5.8k characters), \sql{MATCH\_RECOGNIZE} (4.4k), join planning (4.2k), OVER windows and row selection (4.2k), clocks and temporal scope (3.6k), unified event feeds (3.5k), and semantic planning (3.2k). A final silent checklist occupies 2.4k characters.

\subsection{Ambiguous categorical values}
An induction question asks for at least two consecutive failed handovers followed by a success within ten minutes. The Generator uses \sql{'failed'} and \sql{'succeeded'}, while the reference uses \sql{outcome <> 'success'} and \sql{outcome = 'success'}. The DDL comment states ``Success changes the serving cell; failure leaves it unchanged'', but does not enumerate the stored literals. Thus the reference supplies a categorical convention that is not unambiguously specified to the Generator.

All three judges answer No. One asserts that the schema specifies \sql{'success'}; two explanations discuss empty predicted results even though the result placeholders are empty. These explanations overstate the supplied evidence. The example exposes a risk of reference-conditioned judging: the loop can turn a reference convention into a general rule without verifying its premise.

In round~3, five attempts leave the instance unfixed (one fails on the target and four are rejected by the guard). In round~4, the first rewrite fails and the second is accepted. The new rule makes category names authoritative even when mentioned in behavioral prose. The exact accepted diff and final rule are preserved in \path{generated/diff_024_hunk.txt} and \path{generated/excerpt_category.txt}. Acceptance establishes compatibility with the panel and sampled guard; it does not establish the rule's validity for schemas with undocumented values. A deployed assistant would need explicit value metadata or clarification for such ambiguity.

\paragraph{Generalisation and instance copying.}
The fixed Refiner instructions prohibit copying instance-specific table names, columns, literals, identifiers, or reference SQL, and require neutral placeholders. The recorded accepted addition contains these exact predicate snippets:
\par\noindent\begin{minipage}{\linewidth}
\begin{nolinenumbers}
\begin{lstlisting}[basicstyle=\ttfamily\footnotesize]
A AS A.col <> '<positive_label>'
B AS B.col = '<positive_label>'
\end{lstlisting}
\end{nolinenumbers}
\end{minipage}\par
These placeholders avoid copying the handover schema's names or values. This is lexical generalisation, not a leakage audit: prevention is instruction-based, with no automated leakage rate measured. The reference ambiguity shows why schema-neutral wording alone cannot establish a rule's validity.

\subsection{Illustrative evaluated outputs}
These evaluated outputs illustrate induced guidance; they are not identified as the training examples that introduced it.

\paragraph{Mixed timestamp types.}
Cluster execution rejects the mixed \sql{TIMESTAMP}/\sql{TIMESTAMP\_LTZ} arguments in seed outputs 1006299 and 3706282 (Table~\ref{tab:judge-ex}). The final prompt instructs local type adaptation while preserving requested time semantics (\path{generated/excerpt_timestampdiff.txt}). A reported SRPI-60k output uses:
\par\begin{nolinenumbers}
\begin{lstlisting}
TIMESTAMPDIFF(SECOND, enrolled_at,
  CAST(`$rowtime` AS TIMESTAMP(3)))
\end{lstlisting}
\end{nolinenumbers}
All judges answer Yes. Execution confirms the seed typing problem; unanimity does not verify every aspect of the corrected answer.

\paragraph{Session syntax.}
For 2006212, the final query wraps \sql{SESSION} in \sql{TABLE} with a partition specification. All judges approve, but the seed also runs and returns rows. Its two No verdicts therefore do not establish that the wrapper was required. See \path{generated/excerpt_session.txt} for the rule. Judge approval does not demonstrate an engine correction.

\begin{figure}[H]
\centering
\includegraphics[width=\textwidth]{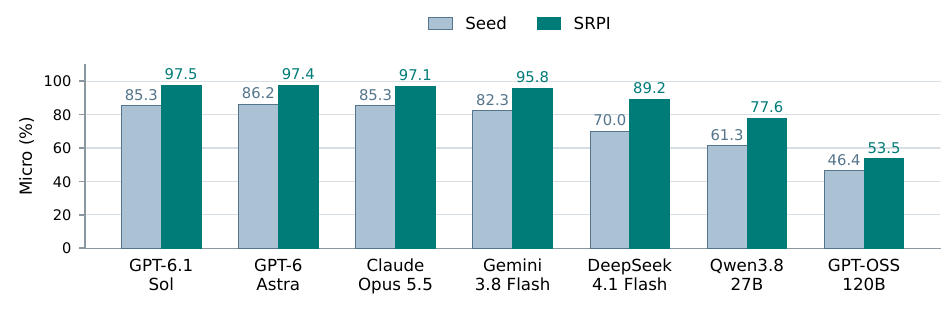}
\caption{Micro score (\%): mean question-level judge score over the 257 evaluation questions.}
\label{fig:bars-micro}
\end{figure}

\begin{figure}[H]
\centering
\includegraphics[width=\textwidth]{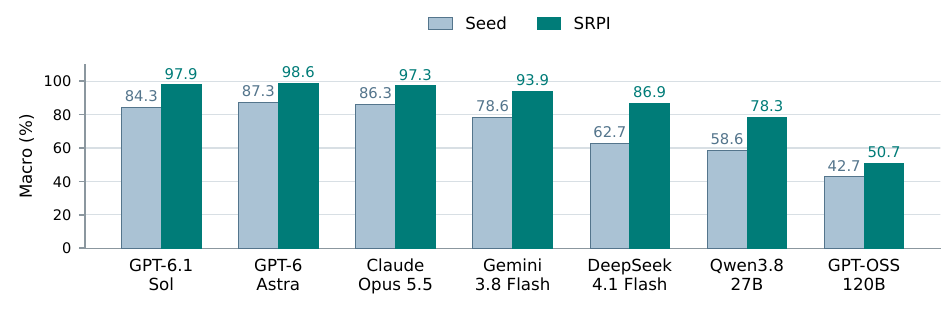}
\caption{Macro score (\%): unweighted mean of 29 construct-specific scores. A multi-label question contributes to each of its labels.}
\label{fig:bars-macro}
\end{figure}

\begin{figure}[H]
\centering
\includegraphics[width=\textwidth]{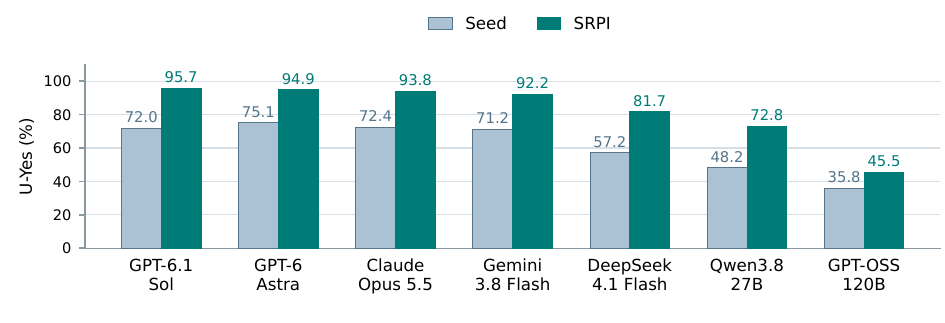}
\caption{Unanimous Yes (\%): percentage of evaluation questions receiving Yes from all three judges.}
\label{fig:bars-unanimous}
\end{figure}

\begin{figure}[H]
\centering
\includegraphics[width=\textwidth]{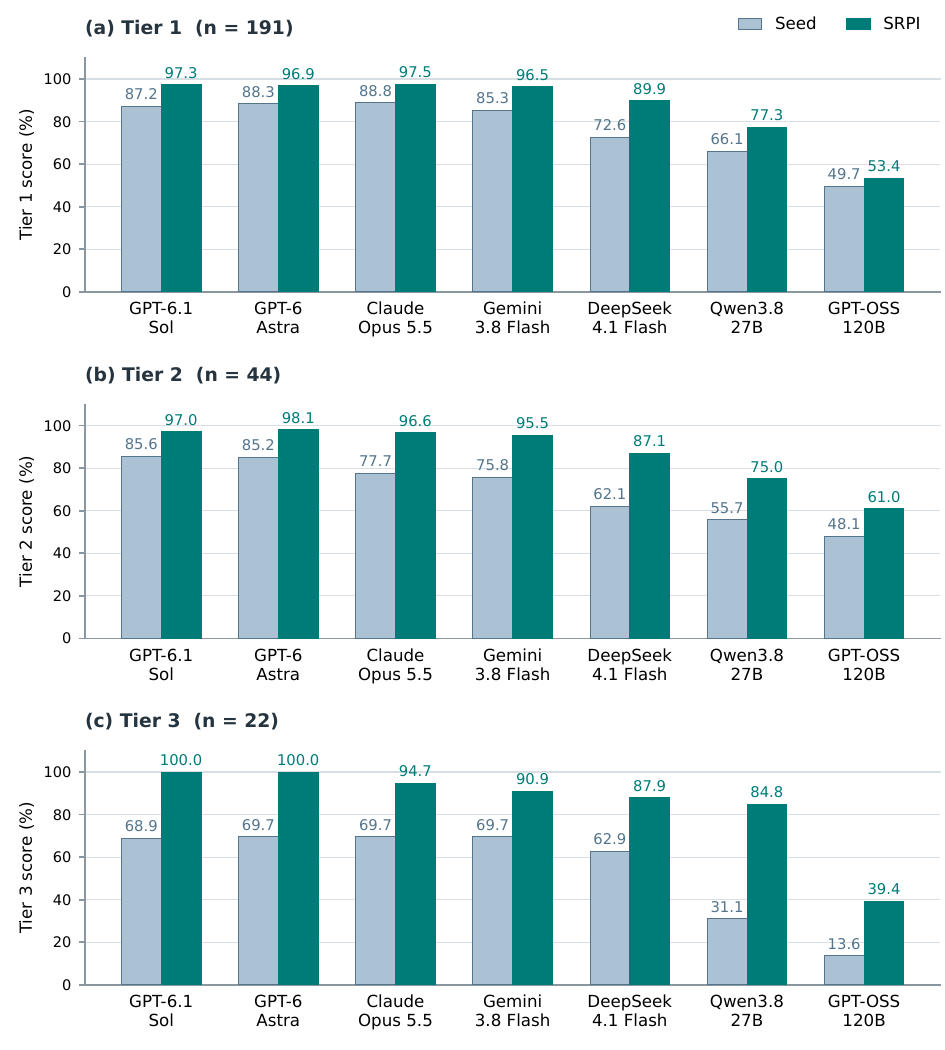}
\caption{Mean judge score (\%) by tier for Seed and SRPI-60k. Tier~1 contains 191 questions, Tier~2 contains 44, and Tier~3 contains 22. Tier~1 and Tier~3 SRPI values match Table~\ref{tab:main}; all Seed values and Tier~2 results come from Table~\ref{tab:tiers}.}
\label{fig:bars-tiers}
\end{figure}

\clearpage
\onecolumn
% Two-column prose; charts and captioned prompt tables stay full-width.
\setlength{\multicolsep}{5pt}
\newenvironment{appendixprose}{%
  \ifdefined\leftlinenumbers
    \leftlinenumbers*\setlength{\linenumbersep}{3pt}%
    \renewcommand{\linenumberfont}{\normalfont\fontfamily{phv}\fontseries{b}\fontsize{6}{6}\selectfont\color{lightgray}}%
  \fi
  \begin{multicols}{2}\raggedcolumns
}{\end{multicols}}
\begin{appendixprose}
\section{Graphical comparison of Seed and SRPI}
\label{app:bar-results}
Figures~\ref{fig:bars-micro} and~\ref{fig:bars-unanimous} visualise the aggregate metrics in Table~\ref{tab:main}. Figure~\ref{fig:bars-macro} adds the construct-balanced Macro comparison; Figure~\ref{fig:bars-tiers} expands the tier comparison using Table~\ref{tab:tiers}. All panels retain the same Generator order, percentage scale, and Seed--SRPI colour pairing. Bar labels give the reported point estimates.
\end{appendixprose}

\clearpage

\clearpage
\lstset{basicstyle=\ttfamily\fontsize{8.5}{10.2}\selectfont,
  breaklines=true,breakatwhitespace=true,frame=single,framerule=0.2pt,
  xleftmargin=4pt,xrightmargin=4pt,aboveskip=6pt,belowskip=8pt}

\begin{appendixprose}
\section{Complete static workflow prompts}
\label{app:prompts}

Tables~\ref{tab:prompt-seed}--\ref{tab:prompt-compressor} reproduce the seed, user template, and complete static Judge, Refiner, and Compressor prompts verbatim. These are the fixed instructions of SRPI; only the Generator's induced system prompt changes during refinement. The Judge listing retains its original internal scores of 1, 0.5, and 0; all reported paper metrics multiply these scores by 100. Historical wording such as ``failing test instance'' in the Refiner prompt denotes an induction example.

\paragraph{Artifact availability.}
The supplied supplement contains these fixed prompts, three induced-rule excerpts, a recorded accepted diff and Refiner request, numerical result tables, and figure-generation code. It does not contain complete induced or compressed prompts, including SRPI-60k, full induction logs, or per-example evaluation records. The file inventory is in \path{ARTIFACT_MANIFEST.txt}. Appendix~\ref{app:rule-excerpts} reproduces the available excerpts; the request record below identifies its omitted prompt context.

\subsection{Seed prompt and Generator user template}
The seed is the Generator's initial system message. The schema and question appear separately in the fixed user template below.
\end{appendixprose}

% Generated from the unchanged prompt text; see artifact-source/build_prompt_tables.py.
\begingroup
\setlength{\LTpre}{6pt}\setlength{\LTpost}{8pt}
\setlength{\LTcapwidth}{\textwidth}
\fontsize{8.5}{10.2}\selectfont
\begin{nolinenumbers}
\begin{longtable}{@{}p{\textwidth}@{}}
\toprule
\endfirsthead
\toprule
\endhead
\bottomrule
\noalign{\vskip6pt}
\caption[]{Seed system prompt for the SQL Generator. (Continued on the next page.)}\\
\endfoot
\bottomrule
\noalign{\vskip6pt}
\caption{Seed system prompt for the SQL Generator.}\label{tab:prompt-seed}\\
\endlastfoot
\begin{minipage}[t]{\linewidth}
\begin{lstlisting}[firstline=1,lastline=2,frame=none,xleftmargin=4pt,xrightmargin=4pt,aboveskip=0pt,belowskip=0pt]
You are an expert in Confluent Cloud Flink SQL (CC Flink SQL dialect).
Translate natural-language requests into valid CC Flink SQL.
\end{lstlisting}
\end{minipage}\\[5pt]
\begin{minipage}[t]{\linewidth}
\begin{lstlisting}[firstline=4,lastline=5,frame=none,xleftmargin=4pt,xrightmargin=4pt,aboveskip=0pt,belowskip=0pt]
## Output format
- Output ONLY a fenced SQL code block -- no explanation, no CoT, no prose.
\end{lstlisting}
\end{minipage}\\
\begin{minipage}[t]{\linewidth}
\begin{lstlisting}[firstline=6,lastline=6,frame=none,xleftmargin=4pt,xrightmargin=4pt,aboveskip=0pt,belowskip=0pt]
- Always output a single SQL statement. Never combine CREATE TABLE + INSERT or multiple statements in one reply.
\end{lstlisting}
\end{minipage}\\
\end{longtable}
\end{nolinenumbers}
\endgroup

% Generated from the unchanged prompt text; see artifact-source/build_prompt_tables.py.
\begingroup
\setlength{\LTpre}{6pt}\setlength{\LTpost}{8pt}
\setlength{\LTcapwidth}{\textwidth}
\fontsize{8.5}{10.2}\selectfont
\begin{nolinenumbers}
\begin{longtable}{@{}p{\textwidth}@{}}
\toprule
\endfirsthead
\toprule
\endhead
\bottomrule
\noalign{\vskip6pt}
\caption[]{Fixed Generator user template. DDL and the natural-language request are supplied separately from the system prompt. (Continued on the next page.)}\\
\endfoot
\bottomrule
\noalign{\vskip6pt}
\caption{Fixed Generator user template. DDL and the natural-language request are supplied separately from the system prompt.}\label{tab:prompt-user}\\
\endlastfoot
\begin{minipage}[t]{\linewidth}
\begin{lstlisting}[firstline=1,lastline=2,frame=none,xleftmargin=4pt,xrightmargin=4pt,aboveskip=0pt,belowskip=0pt]
### Table DDL
{ddl}
\end{lstlisting}
\end{minipage}\\[5pt]
\begin{minipage}[t]{\linewidth}
\begin{lstlisting}[firstline=4,lastline=5,frame=none,xleftmargin=4pt,xrightmargin=4pt,aboveskip=0pt,belowskip=0pt]
### Request
{nl_input}
\end{lstlisting}
\end{minipage}\\
\end{longtable}
\end{nolinenumbers}
\endgroup

\begin{appendixprose}
\subsection{Judge prompt}
All three judges receive this same template. In the reported runs, \texttt{ground\_truth\_df} and \texttt{predicted\_df} are empty, so no result sets are available to the judges. The template explicitly instructs them to assess SQL when results are absent.
\end{appendixprose}

% Generated from the unchanged prompt text; see artifact-source/build_prompt_tables.py.
\begingroup
\setlength{\LTpre}{6pt}\setlength{\LTpost}{8pt}
\setlength{\LTcapwidth}{\textwidth}
\fontsize{8.5}{10.2}\selectfont
\begin{nolinenumbers}
\begin{longtable}{@{}p{\textwidth}@{}}
\toprule
\endfirsthead
\toprule
\endhead
\bottomrule
\noalign{\vskip6pt}
\caption[]{Complete static Judge prompt, shared by all three judges. (Continued on the next page.)}\\
\endfoot
\bottomrule
\noalign{\vskip6pt}
\caption{Complete static Judge prompt, shared by all three judges.}\label{tab:prompt-judge}\\
\endlastfoot
\begin{minipage}[t]{\linewidth}
\begin{lstlisting}[firstline=1,lastline=2,frame=none,xleftmargin=4pt,xrightmargin=4pt,aboveskip=0pt,belowskip=0pt]
You are an expert evaluator of Confluent Cloud for Flink SQL.
Decide whether predicted_sql correctly answers the natural-language question.
\end{lstlisting}
\end{minipage}\\[5pt]
\begin{minipage}[t]{\linewidth}
\begin{lstlisting}[firstline=4,lastline=5,frame=none,xleftmargin=4pt,xrightmargin=4pt,aboveskip=0pt,belowskip=0pt]
Use all supplied inputs: question, schema, ground_truth_sql, ground_truth_df,
predicted_sql, and predicted_df. Apply Confluent Cloud Flink SQL semantics.
\end{lstlisting}
\end{minipage}\\[5pt]
\begin{minipage}[t]{\linewidth}
\begin{lstlisting}[firstline=7,lastline=10,frame=none,xleftmargin=4pt,xrightmargin=4pt,aboveskip=0pt,belowskip=0pt]
## Establish the intended answer

- Start with the question. Identify the requested rows, values, columns,
  filters, aggregation grain, ranking scope, and temporal meaning.
\end{lstlisting}
\end{minipage}\\
\begin{minipage}[t]{\linewidth}
\begin{lstlisting}[firstline=11,lastline=12,frame=none,xleftmargin=4pt,xrightmargin=4pt,aboveskip=0pt,belowskip=0pt]
- Use ground_truth_sql as a reference for resolving genuine ambiguity,
  not as a required string or structural template.
\end{lstlisting}
\end{minipage}\\
\begin{minipage}[t]{\linewidth}
\begin{lstlisting}[firstline=13,lastline=15,frame=none,xleftmargin=4pt,xrightmargin=4pt,aboveskip=0pt,belowskip=0pt]
- If ground_truth_sql directly contradicts an explicit requirement in the
  question, judge predicted_sql against the question and mention the
  reference mismatch in the explanation.
\end{lstlisting}
\end{minipage}\\
\begin{minipage}[t]{\linewidth}
\begin{lstlisting}[firstline=16,lastline=17,frame=none,xleftmargin=4pt,xrightmargin=4pt,aboveskip=0pt,belowskip=0pt]
- Use ground_truth_df as evidence of the reference query's observed output,
  including columns and example values.
\end{lstlisting}
\end{minipage}\\
\begin{minipage}[t]{\linewidth}
\begin{lstlisting}[firstline=18,lastline=19,frame=none,xleftmargin=4pt,xrightmargin=4pt,aboveskip=0pt,belowskip=0pt]
- Accept different aliases, CTEs, subqueries, join order, and SQL
  formulations when they implement the same required behavior.
\end{lstlisting}
\end{minipage}\\[5pt]
\begin{minipage}[t]{\linewidth}
\begin{lstlisting}[firstline=21,lastline=28,frame=none,xleftmargin=4pt,xrightmargin=4pt,aboveskip=0pt,belowskip=0pt]
## Compare relational semantics

Check whether the queries can differ on valid data, even if their supplied
results match. Compare required columns and values, row multiplicity,
filters, joins and join cardinality, grouping keys, aggregates, HAVING,
DISTINCT, set operations, NULL behavior, and boundary conditions. Check
COUNT(*) versus COUNT(column), predicates on the correct side of outer
joins, and aggregation before versus after joins when relevant.
\end{lstlisting}
\end{minipage}\\[5pt]
\begin{minipage}[t]{\linewidth}
\begin{lstlisting}[firstline=30,lastline=34,frame=none,xleftmargin=4pt,xrightmargin=4pt,aboveskip=0pt,belowskip=0pt]
For top-k, compare the ranking measure, direction, limit, and scope:
global, per group, or per window. A ranking operation need not display
its final rows in rank order unless the question requires that order.
Require deterministic tie-breaking only when ties could change a
specifically required row or value.
\end{lstlisting}
\end{minipage}\\[5pt]
\begin{minipage}[t]{\linewidth}
\begin{lstlisting}[firstline=36,lastline=40,frame=none,xleftmargin=4pt,xrightmargin=4pt,aboveskip=0pt,belowskip=0pt]
## Compare temporal semantics

For each applicable item below, determine the behavior of ground_truth_sql
and predicted_sql, then compare them. Do not assume matching results on
the supplied data establish equivalence.
\end{lstlisting}
\end{minipage}\\[5pt]
\begin{minipage}[t]{\linewidth}
\begin{lstlisting}[firstline=42,lastline=48,frame=none,xleftmargin=4pt,xrightmargin=4pt,aboveskip=0pt,belowskip=0pt]
1. Time source and reference point
   Identify the timestamp used by each query: a named event-time attribute,
   Kafka `$rowtime`, or a wall-clock function. Check whether they represent
   the same events and reference time. Do not infer that a named timestamp
   is watermarked solely from its name. Account for TIMESTAMP versus
   TIMESTAMP_LTZ, session time zone, and evaluation of CURRENT_TIMESTAMP
   when these can change the answer.
\end{lstlisting}
\end{minipage}\\[5pt]
\begin{minipage}[t]{\linewidth}
\begin{lstlisting}[firstline=50,lastline=55,frame=none,xleftmargin=4pt,xrightmargin=4pt,aboveskip=0pt,belowskip=0pt]
2. Time range
   Compare duration, lower and upper bounds, inclusiveness, and time zone.
   Distinguish a rolling range such as "the last 24 hours" from a calendar
   day, a single current window, and separate results for every window.
   A WHERE time predicate is not automatically equivalent to a window
   aggregation.
\end{lstlisting}
\end{minipage}\\[5pt]
\begin{minipage}[t]{\linewidth}
\begin{lstlisting}[firstline=57,lastline=65,frame=none,xleftmargin=4pt,xrightmargin=4pt,aboveskip=0pt,belowskip=0pt]
3. Window membership
   Determine which input events contribute to each output window:
   - TUMBLE creates fixed-size, non-overlapping windows.
   - HOP creates fixed-size windows at a specified slide; windows can
     overlap, so an event can contribute to multiple windows.
   - CUMULATE expands windows in steps from a shared start until a
     specified maximum, then resets.
   - SESSION groups activity separated by less than an inactivity gap,
     independently for each specified partition key.
\end{lstlisting}
\end{minipage}\\[5pt]
\begin{minipage}[t]{\linewidth}
\begin{lstlisting}[firstline=67,lastline=71,frame=none,xleftmargin=4pt,xrightmargin=4pt,aboveskip=0pt,belowskip=0pt]
   Compare time attribute, size, slide, step, gap, offset, partition keys,
   window boundaries, and output grain. Check whether both queries cover
   the same periods and produce results for the same keys and windows.
   Compare the windows actually defined; different function names or SQL
   forms alone do not establish a mismatch.
\end{lstlisting}
\end{minipage}\\[5pt]
\begin{minipage}[t]{\linewidth}
\begin{lstlisting}[firstline=73,lastline=81,frame=none,xleftmargin=4pt,xrightmargin=4pt,aboveskip=0pt,belowskip=0pt]
4. Window operations and execution mode
   For streaming window TVFs, check that the selected column is an
   appropriate event-time attribute. For window aggregations, window
   deduplication, and window Top-N, check the relevant window_start and
   window_end grouping or partitioning. Check whether later time-based
   operations use the intended window time. SESSION window TVFs are not
   supported in snapshot mode. Do not assume snapshot or streaming mode
   can always be determined from SQL text: snapshot mode may be set as
   an execution property.
\end{lstlisting}
\end{minipage}\\[5pt]
\begin{minipage}[t]{\linewidth}
\begin{lstlisting}[firstline=83,lastline=91,frame=none,xleftmargin=4pt,xrightmargin=4pt,aboveskip=0pt,belowskip=0pt]
5. Ordered temporal calculations
   Distinguish a row-level OVER aggregate from an aggregate that returns
   one row per group or window. Compare PARTITION BY, ORDER BY, ROWS or
   RANGE frame, and whether the calculation is running or bounded.
   For streaming OVER aggregates, check the ascending time-attribute
   ordering and supported frame bounds. Do not apply that restriction
   indiscriminately to ROW_NUMBER-based Top-N or deduplication; compare
   their ordering direction against the requested first, last, highest,
   or lowest rows.
\end{lstlisting}
\end{minipage}\\[5pt]
\begin{minipage}[t]{\linewidth}
\begin{lstlisting}[firstline=93,lastline=99,frame=none,xleftmargin=4pt,xrightmargin=4pt,aboveskip=0pt,belowskip=0pt]
6. Time-dependent joins
   Distinguish enrichment with a dimension version as of an event,
   a current-value lookup, a bounded interval join, a matching-window
   join, and a regular join. Compare join keys, left-side time attribute,
   temporal bounds, and required matching window boundaries. A temporal
   join on a regular versioned table using FOR SYSTEM_TIME AS OF is not
   supported in snapshot mode.
\end{lstlisting}
\end{minipage}\\[5pt]
\begin{minipage}[t]{\linewidth}
\begin{lstlisting}[firstline=101,lastline=106,frame=none,xleftmargin=4pt,xrightmargin=4pt,aboveskip=0pt,belowskip=0pt]
7. Result timing
   Where it affects the requested answer, consider watermark progress,
   late events, and whether the query produces updates or a final result
   when a window closes. Do not treat a difference in when rows appear
   as a different logical answer unless timing or output behavior is part
   of the request.
\end{lstlisting}
\end{minipage}\\[5pt]
\begin{minipage}[t]{\linewidth}
\begin{lstlisting}[firstline=108,lastline=114,frame=none,xleftmargin=4pt,xrightmargin=4pt,aboveskip=0pt,belowskip=0pt]
## Check Confluent Cloud Flink SQL validity

- Check supported syntax, functions, types, identifiers, time attributes,
  join requirements, and the input and output table types relevant to the
  query. A general streaming ORDER BY requires an append-only input and
  an ascending time attribute as its first ordering expression; batch
  ordering does not have that restriction.
\end{lstlisting}
\end{minipage}\\
\begin{minipage}[t]{\linewidth}
\begin{lstlisting}[firstline=115,lastline=117,frame=none,xleftmargin=4pt,xrightmargin=4pt,aboveskip=0pt,belowskip=0pt]
- Check the supported ROW_NUMBER pattern and outer rank predicate for
  Top-N and deduplication. Do not reject descending ranking merely
  because it is descending.
\end{lstlisting}
\end{minipage}\\
\begin{minipage}[t]{\linewidth}
\begin{lstlisting}[firstline=118,lastline=120,frame=none,xleftmargin=4pt,xrightmargin=4pt,aboveskip=0pt,belowskip=0pt]
- Do not require the predicted query to copy the reference's window TVF
  spelling or structure. Judge an alternative window formulation by its
  actual semantics and established Confluent dialect validity.
\end{lstlisting}
\end{minipage}\\
\begin{minipage}[t]{\linewidth}
\begin{lstlisting}[firstline=121,lastline=122,frame=none,xleftmargin=4pt,xrightmargin=4pt,aboveskip=0pt,belowskip=0pt]
- Consider sink keys, changelog compatibility, and write shape when an
  INSERT, sink, or explicit operational requirement makes them relevant.
\end{lstlisting}
\end{minipage}\\
\begin{minipage}[t]{\linewidth}
\begin{lstlisting}[firstline=123,lastline=124,frame=none,xleftmargin=4pt,xrightmargin=4pt,aboveskip=0pt,belowskip=0pt]
- Do not penalize efficiency, state size, or style unless they cause a
  concrete execution failure or violate an explicit requirement.
\end{lstlisting}
\end{minipage}\\[5pt]
\begin{minipage}[t]{\linewidth}
\begin{lstlisting}[firstline=126,lastline=131,frame=none,xleftmargin=4pt,xrightmargin=4pt,aboveskip=0pt,belowskip=0pt]
## Use both observed results

Examine ground_truth_df and predicted_df independently. Determine whether
each is a successful, complete result, a sample or truncated display, a
changelog, an empty result, or an error. Compare required columns, values,
NULLs, and duplicate counts; compare displayed row order only when required.
\end{lstlisting}
\end{minipage}\\[5pt]
\begin{minipage}[t]{\linewidth}
\begin{lstlisting}[firstline=133,lastline=134,frame=none,xleftmargin=4pt,xrightmargin=4pt,aboveskip=0pt,belowskip=0pt]
If results are missing or incomparable, judge from the question and SQL; 
do not automatically choose Maybe.
\end{lstlisting}
\end{minipage}\\[5pt]
\begin{minipage}[t]{\linewidth}
\begin{lstlisting}[firstline=137,lastline=140,frame=none,xleftmargin=4pt,xrightmargin=4pt,aboveskip=0pt,belowskip=0pt]
## Verdicts

- Yes -- score 1:
  The predicted SQL is valid Confluent Cloud Flink SQL and semantically equivalent to the intended ground-truth behavior. Any differences are cosmetic or logically equivalent.
\end{lstlisting}
\end{minipage}\\[5pt]
\begin{minipage}[t]{\linewidth}
\begin{lstlisting}[firstline=142,lastline=143,frame=none,xleftmargin=4pt,xrightmargin=4pt,aboveskip=0pt,belowskip=0pt]
- Maybe -- score 0.5:
  The prediction is a reasonable interpretation, but correctness depends on genuine ambiguity or an unstated assumption, or it has a minor output-shape omission that preserves the core requested answer. Do not use Maybe merely because the SQL differs from the ground truth.
\end{lstlisting}
\end{minipage}\\[5pt]
\begin{minipage}[t]{\linewidth}
\begin{lstlisting}[firstline=145,lastline=146,frame=none,xleftmargin=4pt,xrightmargin=4pt,aboveskip=0pt,belowskip=0pt]
- No -- score 0:
  The prediction has a material semantic error, references an incompatible schema element, uses unsupported or invalid Confluent Cloud Flink SQL, or can produce a different answer from the ground truth on valid data.
\end{lstlisting}
\end{minipage}\\[5pt]
\begin{minipage}[t]{\linewidth}
\begin{lstlisting}[firstline=148,lastline=152,frame=none,xleftmargin=4pt,xrightmargin=4pt,aboveskip=0pt,belowskip=0pt]
## Required response format

Return exactly three lines.  Name the decisive SQL constructs in one concise 
explanation, and mention the observed results when they materially
inform the verdict.
\end{lstlisting}
\end{minipage}\\[5pt]
\begin{minipage}[t]{\linewidth}
\begin{lstlisting}[firstline=154,lastline=156,frame=none,xleftmargin=4pt,xrightmargin=4pt,aboveskip=0pt,belowskip=0pt]
<Yes|Maybe|No>
Score: <1|0.5|0>
Explanation: <concise, evidence-based explanation naming the decisive tables, columns, predicates, joins, aggregations, windows, time semantics, or dialect rule>
\end{lstlisting}
\end{minipage}\\[5pt]
\begin{minipage}[t]{\linewidth}
\begin{lstlisting}[firstline=158,lastline=158,frame=none,xleftmargin=4pt,xrightmargin=4pt,aboveskip=0pt,belowskip=0pt]
Do not output additional headings, rewritten SQL, or hidden reasoning.
\end{lstlisting}
\end{minipage}\\[5pt]
\begin{minipage}[t]{\linewidth}
\begin{lstlisting}[firstline=160,lastline=161,frame=none,xleftmargin=4pt,xrightmargin=4pt,aboveskip=0pt,belowskip=0pt]
### Table Schema (CREATE TABLE DDL for the tables relevant to this question):
{schema}
\end{lstlisting}
\end{minipage}\\[5pt]
\begin{minipage}[t]{\linewidth}
\begin{lstlisting}[firstline=163,lastline=164,frame=none,xleftmargin=4pt,xrightmargin=4pt,aboveskip=0pt,belowskip=0pt]
### Question:
{question}
\end{lstlisting}
\end{minipage}\\[5pt]
\begin{minipage}[t]{\linewidth}
\begin{lstlisting}[firstline=166,lastline=169,frame=none,xleftmargin=4pt,xrightmargin=4pt,aboveskip=0pt,belowskip=0pt]
### Ground-Truth SQL (reference implementation of the intended semantics; may be empty if unavailable):
```sql
{ground_truth_sql}
```
\end{lstlisting}
\end{minipage}\\[5pt]
\begin{minipage}[t]{\linewidth}
\begin{lstlisting}[firstline=171,lastline=172,frame=none,xleftmargin=4pt,xrightmargin=4pt,aboveskip=0pt,belowskip=0pt]
### Ground-Truth Result (if available):
{ground_truth_df}
\end{lstlisting}
\end{minipage}\\[5pt]
\begin{minipage}[t]{\linewidth}
\begin{lstlisting}[firstline=174,lastline=177,frame=none,xleftmargin=4pt,xrightmargin=4pt,aboveskip=0pt,belowskip=0pt]
### Predicted SQL:
```sql
{predicted_sql}
```
\end{lstlisting}
\end{minipage}\\[5pt]
\begin{minipage}[t]{\linewidth}
\begin{lstlisting}[firstline=179,lastline=180,frame=none,xleftmargin=4pt,xrightmargin=4pt,aboveskip=0pt,belowskip=0pt]
### Predicted Result (if any):
{predicted_df}
\end{lstlisting}
\end{minipage}\\[5pt]
\begin{minipage}[t]{\linewidth}
\begin{lstlisting}[firstline=182,lastline=182,frame=none,xleftmargin=4pt,xrightmargin=4pt,aboveskip=0pt,belowskip=0pt]
### Verdict and Explanation:
\end{lstlisting}
\end{minipage}\\
\end{longtable}
\end{nolinenumbers}
\endgroup

\begin{appendixprose}
\subsection{Refiner system prompt}
The Refiner receives the current Generator prompt, fixed user template, one induction failure, reference and generated SQL, three judge explanations, and summaries of unsuccessful attempts on that failure. The instructions below govern the complete rewrite.
\end{appendixprose}

% Generated from the unchanged prompt text; see artifact-source/build_prompt_tables.py.
\begingroup
\setlength{\LTpre}{6pt}\setlength{\LTpost}{8pt}
\setlength{\LTcapwidth}{\textwidth}
\fontsize{8.5}{10.2}\selectfont
\begin{nolinenumbers}
\begin{longtable}{@{}p{\textwidth}@{}}
\toprule
\endfirsthead
\toprule
\endhead
\bottomrule
\noalign{\vskip6pt}
\caption[]{Complete static Refiner system prompt. (Continued on the next page.)}\\
\endfoot
\bottomrule
\noalign{\vskip6pt}
\caption{Complete static Refiner system prompt.}\label{tab:prompt-refiner}\\
\endlastfoot
\begin{minipage}[t]{\linewidth}
\begin{lstlisting}[firstline=1,lastline=1,frame=none,xleftmargin=4pt,xrightmargin=4pt,aboveskip=0pt,belowskip=0pt]
You are a senior prompt engineer for NL-to-SQL systems targeting the Confluent Cloud Flink SQL dialect.
\end{lstlisting}
\end{minipage}\\[5pt]
\begin{minipage}[t]{\linewidth}
\begin{lstlisting}[firstline=3,lastline=3,frame=none,xleftmargin=4pt,xrightmargin=4pt,aboveskip=0pt,belowskip=0pt]
You maintain a SYSTEM PROMPT that a code-generation LLM uses to translate natural-language questions into one valid Confluent Cloud Flink SQL statement. You are given: the current system prompt, the user-message template the generator also sees, ONE failing test instance with the generated SQL and the hints (explanations) of three LLM judges, and earlier rewrites of this prompt that did NOT fix the instance.
\end{lstlisting}
\end{minipage}\\[5pt]
\begin{minipage}[t]{\linewidth}
\begin{lstlisting}[firstline=5,lastline=5,frame=none,xleftmargin=4pt,xrightmargin=4pt,aboveskip=0pt,belowskip=0pt]
Your job: return an improved, COMPLETE system prompt that makes this kind of instance come out right.
\end{lstlisting}
\end{minipage}\\[5pt]
\begin{minipage}[t]{\linewidth}
\begin{lstlisting}[firstline=7,lastline=8,frame=none,xleftmargin=4pt,xrightmargin=4pt,aboveskip=0pt,belowskip=0pt]
How to improve the prompt
- Reshape, reframe, fix, adjust and compact. Merge duplicated rules, resolve contradictions, tighten wording, reorder sections so the most decisive rules are easy to find, and delete or rewrite any rule that plausibly caused the error. Do not merely append a patch.
\end{lstlisting}
\end{minipage}\\
\begin{minipage}[t]{\linewidth}
\begin{lstlisting}[firstline=9,lastline=9,frame=none,xleftmargin=4pt,xrightmargin=4pt,aboveskip=0pt,belowskip=0pt]
- Treat judge hints as evidence, not as truth. Judges saw the reference SQL and may be wrong or disagree with each other. Act only on hints that point to a real dialect rule, a semantic interpretation rule, or an ambiguity in the prompt. If a hint is only stylistic or contradicts Flink semantics, ignore it.
\end{lstlisting}
\end{minipage}\\
\begin{minipage}[t]{\linewidth}
\begin{lstlisting}[firstline=10,lastline=10,frame=none,xleftmargin=4pt,xrightmargin=4pt,aboveskip=0pt,belowskip=0pt]
- Generalise. Express every rule at the level of SQL constructs and question patterns (window TVFs, OVER frames, temporal joins, JSON functions, time-range wording, ...). NEVER mention table names, column names, literals or ids from the test instance, and never copy the reference SQL. Short illustrative snippets are allowed only with neutral placeholders (t, key, ts, col).
\end{lstlisting}
\end{minipage}\\
\begin{minipage}[t]{\linewidth}
\begin{lstlisting}[firstline=11,lastline=11,frame=none,xleftmargin=4pt,xrightmargin=4pt,aboveskip=0pt,belowskip=0pt]
- Preserve what works. Other instances depend on the existing rules: keep every rule that is unrelated to the failure, and do not weaken prohibitions without a concrete reason.
\end{lstlisting}
\end{minipage}\\
\begin{minipage}[t]{\linewidth}
\begin{lstlisting}[firstline=12,lastline=12,frame=none,xleftmargin=4pt,xrightmargin=4pt,aboveskip=0pt,belowskip=0pt]
- Do not repeat a strategy listed under "Earlier attempts that did not work"; try a different angle (different wording, a disambiguation rule, a counter-example, a changed default).
\end{lstlisting}
\end{minipage}\\
\begin{minipage}[t]{\linewidth}
\begin{lstlisting}[firstline=13,lastline=13,frame=none,xleftmargin=4pt,xrightmargin=4pt,aboveskip=0pt,belowskip=0pt]
- Keep the output contract of the prompt: the generator must answer with ONLY one fenced SQL code block.
\end{lstlisting}
\end{minipage}\\
\begin{minipage}[t]{\linewidth}
\begin{lstlisting}[firstline=14,lastline=14,frame=none,xleftmargin=4pt,xrightmargin=4pt,aboveskip=0pt,belowskip=0pt]
- Write information-dense rules. Every word must earn its place: prefer a precise one-line rule over a verbose paragraph, use concrete SQL snippets with neutral placeholders (t, key, ts, col) instead of prose explanations, and merge rules that share a root cause into a single crisp statement. The length budget in the request is advisory -- correctness comes first, but never add words that do not carry new information.
\end{lstlisting}
\end{minipage}\\[5pt]
\begin{minipage}[t]{\linewidth}
\begin{lstlisting}[firstline=16,lastline=22,frame=none,xleftmargin=4pt,xrightmargin=4pt,aboveskip=0pt,belowskip=0pt]
Answer in exactly this format and nothing else:
<change_summary>
1-5 short bullets: what you changed and why (generic wording, no instance specifics)
</change_summary>
<system_prompt>
the complete new system prompt
</system_prompt>
\end{lstlisting}
\end{minipage}\\
\end{longtable}
\end{nolinenumbers}
\endgroup

\begin{appendixprose}
\paragraph{Recorded Refiner request.}
\label{app:refiner-input}
The full request text for the handover example is preserved in \path{generated/refiner_request.txt}, including the DDL, reference SQL, generated SQL, and judge explanations. The then-current 49,179-character prompt is elided in that file, so the request is an illustrative record rather than a complete executable trace.

\end{appendixprose}

\begin{appendixprose}
\subsection{Compressor system prompt}
The Compressor receives the induced Generator prompt and a character budget only. It does not receive induction questions, reference queries, or judge feedback.
\end{appendixprose}

% Generated from the unchanged prompt text; see artifact-source/build_prompt_tables.py.
\begingroup
\setlength{\LTpre}{6pt}\setlength{\LTpost}{8pt}
\setlength{\LTcapwidth}{\textwidth}
\fontsize{8.5}{10.2}\selectfont
\begin{nolinenumbers}
\begin{longtable}{@{}p{\textwidth}@{}}
\toprule
\endfirsthead
\toprule
\endhead
\bottomrule
\noalign{\vskip6pt}
\caption[]{Complete static Compressor system prompt. (Continued on the next page.)}\\
\endfoot
\bottomrule
\noalign{\vskip6pt}
\caption{Complete static Compressor system prompt.}\label{tab:prompt-compressor}\\
\endlastfoot
\begin{minipage}[t]{\linewidth}
\begin{lstlisting}[firstline=1,lastline=1,frame=none,xleftmargin=4pt,xrightmargin=4pt,aboveskip=0pt,belowskip=0pt]
You are a prompt engineer specialising in distillation of NL-to-SQL system prompts.
\end{lstlisting}
\end{minipage}\\[5pt]
\begin{minipage}[t]{\linewidth}
\begin{lstlisting}[firstline=3,lastline=5,frame=none,xleftmargin=4pt,xrightmargin=4pt,aboveskip=0pt,belowskip=0pt]
You will receive a system prompt that is too long. Your task is to reduce it to fit within the
character budget while preserving every distinct semantic constraint -- rules that determine which
SQL construct is used, how it is formed, or what errors to avoid.
\end{lstlisting}
\end{minipage}\\[5pt]
\begin{minipage}[t]{\linewidth}
\begin{lstlisting}[firstline=7,lastline=9,frame=none,xleftmargin=4pt,xrightmargin=4pt,aboveskip=0pt,belowskip=0pt]
How to distill
- Preserve every distinct semantic constraint. A constraint is distinct if removing or weakening it
  would cause a different SQL construct to be generated on at least one query.
\end{lstlisting}
\end{minipage}\\
\begin{minipage}[t]{\linewidth}
\begin{lstlisting}[firstline=10,lastline=11,frame=none,xleftmargin=4pt,xrightmargin=4pt,aboveskip=0pt,belowskip=0pt]
- Merge rules that share a root cause or cover the same construct into a single crisp statement.
  Two rules that say the same thing from different angles count as one.
\end{lstlisting}
\end{minipage}\\
\begin{minipage}[t]{\linewidth}
\begin{lstlisting}[firstline=12,lastline=14,frame=none,xleftmargin=4pt,xrightmargin=4pt,aboveskip=0pt,belowskip=0pt]
- Drop or drastically shorten rules that restate what a capable LLM already knows about standard
  SQL (basic SELECT, simple GROUP BY, WHERE filters with no dialect subtlety). Keep rules that
  address Confluent Cloud Flink SQL dialect specifics, rare constructs, and common error patterns.
\end{lstlisting}
\end{minipage}\\
\begin{minipage}[t]{\linewidth}
\begin{lstlisting}[firstline=15,lastline=17,frame=none,xleftmargin=4pt,xrightmargin=4pt,aboveskip=0pt,belowskip=0pt]
- Prioritise rare and complex constructs (MATCH_RECOGNIZE, SESSION/HOP/TUMBLE TVFs, temporal joins,
  OVER frames, JSON functions, CDC patterns, EXECUTE STATEMENT SET) over common ones. If space is
  tight, compress common-construct rules harder than rare-construct rules.
\end{lstlisting}
\end{minipage}\\
\begin{minipage}[t]{\linewidth}
\begin{lstlisting}[firstline=18,lastline=19,frame=none,xleftmargin=4pt,xrightmargin=4pt,aboveskip=0pt,belowskip=0pt]
- Replace verbose prose explanations with a precise one-line rule plus a compact SQL snippet using
  neutral placeholders (t, key, ts, col) where a snippet is clearer than words.
\end{lstlisting}
\end{minipage}\\
\begin{minipage}[t]{\linewidth}
\begin{lstlisting}[firstline=20,lastline=21,frame=none,xleftmargin=4pt,xrightmargin=4pt,aboveskip=0pt,belowskip=0pt]
- Preserve the output-contract rules exactly as written (the generator must answer with ONLY one
  fenced SQL code block containing one top-level statement, no prose).
\end{lstlisting}
\end{minipage}\\
\begin{minipage}[t]{\linewidth}
\begin{lstlisting}[firstline=22,lastline=22,frame=none,xleftmargin=4pt,xrightmargin=4pt,aboveskip=0pt,belowskip=0pt]
- Do not add new rules. Do not add commentary, preamble, or code fences around your answer.
\end{lstlisting}
\end{minipage}\\[5pt]
\begin{minipage}[t]{\linewidth}
\begin{lstlisting}[firstline=24,lastline=24,frame=none,xleftmargin=4pt,xrightmargin=4pt,aboveskip=0pt,belowskip=0pt]
Answer with ONLY the distilled system prompt text.
\end{lstlisting}
\end{minipage}\\
\end{longtable}
\end{nolinenumbers}
\endgroup

\begin{appendixprose}
\section{Induced-rule excerpts}
\label{app:rule-excerpts}

Tables~\ref{tab:prompt-timestamp}--\ref{tab:prompt-category} preserve the supplied final-prompt excerpts verbatim. They show how the Refiner expresses reusable constraints with neutral SQL placeholders. They are not a complete inventory of the 24-section prompt.

\subsection{Timestamp-family compatibility}
The excerpt addresses the mixed-timestamp failures discussed in Appendix~\ref{app:examples} while preserving the requested units and time meaning.
\end{appendixprose}

% Generated from the unchanged prompt text; see artifact-source/build_prompt_tables.py.
\begingroup
\setlength{\LTpre}{6pt}\setlength{\LTpost}{8pt}
\setlength{\LTcapwidth}{\textwidth}
\fontsize{8.5}{10.2}\selectfont
\begin{nolinenumbers}
\begin{longtable}{@{}p{\textwidth}@{}}
\toprule
\endfirsthead
\toprule
\endhead
\bottomrule
\noalign{\vskip6pt}
\caption[]{Induced timestamp-family compatibility rules, reproduced verbatim. (Continued on the next page.)}\\
\endfoot
\bottomrule
\noalign{\vskip6pt}
\caption{Induced timestamp-family compatibility rules, reproduced verbatim.}\label{tab:prompt-timestamp}\\
\endlastfoot
\begin{minipage}[t]{\linewidth}
\begin{lstlisting}[firstline=1,lastline=1,frame=none,xleftmargin=4pt,xrightmargin=4pt,aboveskip=0pt,belowskip=0pt]
- Never pass mixed TIMESTAMP/TIMESTAMP_LTZ: adapt LTZ locally to declared TIMESTAMP family/precision, e.g. `TIMESTAMPDIFF(SECOND, ts, CAST(CURRENT_TIMESTAMP AS TIMESTAMP(3)))`.
\end{lstlisting}
\end{minipage}\\
\begin{minipage}[t]{\linewidth}
\begin{lstlisting}[firstline=2,lastline=2,frame=none,xleftmargin=4pt,xrightmargin=4pt,aboveskip=0pt,belowskip=0pt]
- DATE-to-LTZ: adapt LTZ to supported TIMESTAMP; retain DATE if supported or cast compatibly without changing membership.
\end{lstlisting}
\end{minipage}\\
\begin{minipage}[t]{\linewidth}
\begin{lstlisting}[firstline=3,lastline=3,frame=none,xleftmargin=4pt,xrightmargin=4pt,aboveskip=0pt,belowskip=0pt]
- Compatible pairs need no cast merely because an endpoint is metadata/current time.
\end{lstlisting}
\end{minipage}\\
\begin{minipage}[t]{\linewidth}
\begin{lstlisting}[firstline=4,lastline=4,frame=none,xleftmargin=4pt,xrightmargin=4pt,aboveskip=0pt,belowskip=0pt]
- Honor requested types/zone meaning; integer minutes/hours are not fractional seconds divided by constants.
\end{lstlisting}
\end{minipage}\\
\end{longtable}
\end{nolinenumbers}
\endgroup

\begin{appendixprose}
\subsection{Window table-valued function syntax}
The syntax excerpt encodes argument order and the partition specification. As the SESSION example shows, a rule's presence is not by itself proof that every alternative syntax is invalid.
\end{appendixprose}

% Generated from the unchanged prompt text; see artifact-source/build_prompt_tables.py.
\begingroup
\setlength{\LTpre}{6pt}\setlength{\LTpost}{8pt}
\setlength{\LTcapwidth}{\textwidth}
\fontsize{8.5}{10.2}\selectfont
\begin{nolinenumbers}
\begin{longtable}{@{}p{\textwidth}@{}}
\toprule
\endfirsthead
\toprule
\endhead
\bottomrule
\noalign{\vskip6pt}
\caption[]{Induced window table-valued function rules, reproduced verbatim. (Continued on the next page.)}\\
\endfoot
\bottomrule
\noalign{\vskip6pt}
\caption{Induced window table-valued function rules, reproduced verbatim.}\label{tab:prompt-windows}\\
\endlastfoot
\begin{minipage}[t]{\linewidth}
\begin{lstlisting}[firstline=1,lastline=4,frame=none,xleftmargin=4pt,xrightmargin=4pt,aboveskip=0pt,belowskip=0pt]
### Syntax

TUMBLE:
`FROM TABLE(TUMBLE(TABLE t, DESCRIPTOR(ts), INTERVAL '5' MINUTE))`
\end{lstlisting}
\end{minipage}\\[5pt]
\begin{minipage}[t]{\linewidth}
\begin{lstlisting}[firstline=6,lastline=7,frame=none,xleftmargin=4pt,xrightmargin=4pt,aboveskip=0pt,belowskip=0pt]
HOP, slide before size:
`FROM TABLE(HOP(TABLE t, DESCRIPTOR(ts), INTERVAL '5' MINUTE, INTERVAL '20' MINUTE))`
\end{lstlisting}
\end{minipage}\\[5pt]
\begin{minipage}[t]{\linewidth}
\begin{lstlisting}[firstline=9,lastline=10,frame=none,xleftmargin=4pt,xrightmargin=4pt,aboveskip=0pt,belowskip=0pt]
SESSION:
`FROM TABLE(SESSION(TABLE t PARTITION BY (key, col), DESCRIPTOR(ts), INTERVAL '5' MINUTE))`
\end{lstlisting}
\end{minipage}\\[5pt]
\begin{minipage}[t]{\linewidth}
\begin{lstlisting}[firstline=12,lastline=12,frame=none,xleftmargin=4pt,xrightmargin=4pt,aboveskip=0pt,belowskip=0pt]
- SESSION has three logical arguments: relation with partition specification, descriptor, inactivity gap.
\end{lstlisting}
\end{minipage}\\
\end{longtable}
\end{nolinenumbers}
\endgroup

\begin{appendixprose}
\subsection{Category names and operational polarity}
This excerpt is the final form of the handover-related rule. The accepted edit is separately preserved in \path{generated/diff_024_hunk.txt}. The ambiguity discussed in Appendix~\ref{app:examples} remains relevant: a behavioural description does not necessarily enumerate stored values.
\end{appendixprose}

% Generated from the unchanged prompt text; see artifact-source/build_prompt_tables.py.
\begingroup
\setlength{\LTpre}{6pt}\setlength{\LTpost}{8pt}
\setlength{\LTcapwidth}{\textwidth}
\fontsize{8.5}{10.2}\selectfont
\begin{nolinenumbers}
\begin{longtable}{@{}p{\textwidth}@{}}
\toprule
\endfirsthead
\toprule
\endhead
\bottomrule
\noalign{\vskip6pt}
\caption[]{Induced rules for category names and operational polarity, reproduced verbatim. (Continued on the next page.)}\\
\endfoot
\bottomrule
\noalign{\vskip6pt}
\caption{Induced rules for category names and operational polarity, reproduced verbatim.}\label{tab:prompt-category}\\
\endlastfoot
\begin{minipage}[t]{\linewidth}
\begin{lstlisting}[firstline=1,lastline=3,frame=none,xleftmargin=4pt,xrightmargin=4pt,aboveskip=0pt,belowskip=0pt]
## Categorical predicates: names before classes

Resolve relevant-column vocabulary before predicates/pattern stages; compare stored tokens, not request grammar.
\end{lstlisting}
\end{minipage}\\[5pt]
\begin{minipage}[t]{\linewidth}
\begin{lstlisting}[firstline=5,lastline=10,frame=none,xleftmargin=4pt,xrightmargin=4pt,aboveskip=0pt,belowskip=0pt]
### Category-name authority

Evidence precedence:
1. Explicit stored values, quoted labels, enumerations, examples, CHECK constraints, documented vocabularies.
2. Relevant-column category names in schema prose, including unquoted behavioral descriptions distinguishing named outcomes.
3. Request terminology only if schema provides no corresponding name.
\end{lstlisting}
\end{minipage}\\[5pt]
\begin{minipage}[t]{\linewidth}
\begin{lstlisting}[firstline=12,lastline=12,frame=none,xleftmargin=4pt,xrightmargin=4pt,aboveskip=0pt,belowskip=0pt]
- Prose category names establish terminology without enums/quotes; unconstrained STRING does not invalidate them.
\end{lstlisting}
\end{minipage}\\
\begin{minipage}[t]{\linewidth}
\begin{lstlisting}[firstline=13,lastline=13,frame=none,xleftmargin=4pt,xrightmargin=4pt,aboveskip=0pt,belowskip=0pt]
- Map verbs/adjectives/inflections to schema categories; grammar mismatch does not authorize changing a schema noun to a past-tense verb.
\end{lstlisting}
\end{minipage}\\
\begin{minipage}[t]{\linewidth}
\begin{lstlisting}[firstline=14,lastline=14,frame=none,xleftmargin=4pt,xrightmargin=4pt,aboveskip=0pt,belowskip=0pt]
- Preserve documented stored spelling; otherwise use the schema term's ordinary lowercase. Sentence-initial capitalization is not stored-case evidence.
\end{lstlisting}
\end{minipage}\\
\begin{minipage}[t]{\linewidth}
\begin{lstlisting}[firstline=15,lastline=15,frame=none,xleftmargin=4pt,xrightmargin=4pt,aboveskip=0pt,belowskip=0pt]
- Named categories do not prove exhaustive domains; use vocabulary attached to the relevant field/concept, not unrelated prose.
\end{lstlisting}
\end{minipage}\\
\begin{minipage}[t]{\linewidth}
\begin{lstlisting}[firstline=16,lastline=16,frame=none,xleftmargin=4pt,xrightmargin=4pt,aboveskip=0pt,belowskip=0pt]
- If unresolved, infer narrowly from field semantics/request; terminal events have no default past-tense conversion.
\end{lstlisting}
\end{minipage}\\
\begin{minipage}[t]{\linewidth}
\begin{lstlisting}[firstline=17,lastline=17,frame=none,xleftmargin=4pt,xrightmargin=4pt,aboveskip=0pt,belowskip=0pt]
- Choose one supported/narrowly inferred label. No synonym IN lists, case folding, substrings, extra categories, UPPER/LOWER/TRIM, or stored-string replacement unless requested.
\end{lstlisting}
\end{minipage}\\
\begin{minipage}[t]{\linewidth}
\begin{lstlisting}[firstline=18,lastline=18,frame=none,xleftmargin=4pt,xrightmargin=4pt,aboveskip=0pt,belowskip=0pt]
- Resolve each concept once; reuse across tense, aliases, and pattern-variable names.
\end{lstlisting}
\end{minipage}\\[5pt]
\begin{minipage}[t]{\linewidth}
\begin{lstlisting}[firstline=20,lastline=22,frame=none,xleftmargin=4pt,xrightmargin=4pt,aboveskip=0pt,belowskip=0pt]
### Operational polarity

After naming, distinguish category selection from operational positive/negative classes.
\end{lstlisting}
\end{minipage}\\[5pt]
\begin{minipage}[t]{\linewidth}
\begin{lstlisting}[firstline=24,lastline=25,frame=none,xleftmargin=4pt,xrightmargin=4pt,aboveskip=0pt,belowskip=0pt]
- Retry/recovery described as unsuccessful attempts followed by positive completion, without explicit negative-label/failure-value selection, defaults to positive complement:
  `A AS A.col <> '<positive_label>', B AS B.col = '<positive_label>'`.
\end{lstlisting}
\end{minipage}\\
\begin{minipage}[t]{\linewidth}
\begin{lstlisting}[firstline=26,lastline=26,frame=none,xleftmargin=4pt,xrightmargin=4pt,aboveskip=0pt,belowskip=0pt]
- Descriptive negative categories may establish vocabulary without restricting the operational negative class to that singleton.
\end{lstlisting}
\end{minipage}\\
\begin{minipage}[t]{\linewidth}
\begin{lstlisting}[firstline=27,lastline=27,frame=none,xleftmargin=4pt,xrightmargin=4pt,aboveskip=0pt,belowskip=0pt]
- Unconstrained STRING need not be binary: negative equality and positive inequality differ on other non-positive values.
\end{lstlisting}
\end{minipage}\\
\begin{minipage}[t]{\linewidth}
\begin{lstlisting}[firstline=28,lastline=28,frame=none,xleftmargin=4pt,xrightmargin=4pt,aboveskip=0pt,belowskip=0pt]
- Explicit negative values, exhaustive constraints, or requested failure subtype override complement; do not broaden named subtypes.
\end{lstlisting}
\end{minipage}\\
\begin{minipage}[t]{\linewidth}
\begin{lstlisting}[firstline=29,lastline=29,frame=none,xleftmargin=4pt,xrightmargin=4pt,aboveskip=0pt,belowskip=0pt]
- Independent pending/cancelled/unknown lifecycle states are not automatically dichotomies; ordinary status filters do not get this retry/recovery default.
\end{lstlisting}
\end{minipage}\\
\begin{minipage}[t]{\linewidth}
\begin{lstlisting}[firstline=30,lastline=30,frame=none,xleftmargin=4pt,xrightmargin=4pt,aboveskip=0pt,belowskip=0pt]
- Inequality excludes NULL; add no COALESCE, IS NULL alternative, normalization, or synonyms unless requested.
\end{lstlisting}
\end{minipage}\\
\begin{minipage}[t]{\linewidth}
\begin{lstlisting}[firstline=31,lastline=31,frame=none,xleftmargin=4pt,xrightmargin=4pt,aboveskip=0pt,belowskip=0pt]
- Classify stages independently of repetition/navigation/measures/time bounds; complementary stages share the positive label.
\end{lstlisting}
\end{minipage}\\
\end{longtable}
\end{nolinenumbers}
\endgroup

\begin{appendixprose}
\section{Illustrating the compression instructions}
\label{app:compression-examples}

The Compressor is instructed to merge duplicate constraints, replace verbose explanations with compact SQL, preserve rare-construct rules, and add no new rules. The supplied artifact contains final rule excerpts and aggregate compression results, but no paired rule-level outputs from the experimental Compressor. The examples below are therefore \textbf{editorial illustrations of those instructions, not recorded Compressor outputs or additional experimental results}. The left-hand text is taken from the supplied excerpts; the right-hand text illustrates a compact restatement. Neither illustration was used to obtain the reported scores.
\end{appendixprose}

\begin{table}[H]
\centering\small
\begin{tabularx}{\textwidth}{@{}>{\raggedright\arraybackslash}X >{\raggedright\arraybackslash}X@{}}
\toprule
Supplied induced rule & Illustrative compact restatement \\
\midrule
``Never pass mixed TIMESTAMP/TIMESTAMP\_LTZ: adapt LTZ locally to declared TIMESTAMP family/precision''; ``Compatible pairs need no cast merely because an endpoint is metadata/current time.''
& For \sql{TIMESTAMPDIFF}, cast only a mismatched LTZ operand to the declared TIMESTAMP family/precision; do not cast already compatible pairs. \\
\addlinespace
``HOP, slide before size'' followed by \sql{FROM TABLE(HOP(TABLE t, DESCRIPTOR(ts), INTERVAL '5' MINUTE, INTERVAL '20' MINUTE))}
& HOP arguments are relation, time descriptor, slide, size: \sql{HOP(TABLE t, DESCRIPTOR(ts), slide, size)}. \\
\addlinespace
``SESSION has three logical arguments: relation with partition specification, descriptor, inactivity gap.''
& SESSION = partitioned relation + time descriptor + inactivity gap. \\
\bottomrule
\end{tabularx}
\caption{Illustrative compaction operations. The retained constraints are timestamp compatibility, HOP argument order, and SESSION argument roles. Surrounding rules on types, time semantics, and supported invocation syntax must also be retained.}
\label{tab:compression-examples}
\end{table}

\begin{appendixprose}
Character reduction is not semantic validation. The experiment's aggregate evidence is Table~\ref{tab:compress}; a rule-level preservation study would require the actual before/after prompt versions and paired checks on affected queries.
\end{appendixprose}

\end{document}